\documentclass[review]{article}
\usepackage{graphicx}
\usepackage{amsfonts}
\usepackage{caption}
\usepackage{subcaption}
\usepackage{booktabs}  
\usepackage{amsmath}
\usepackage{amsthm}
\usepackage{caption}
\usepackage{textcomp}  
\usepackage{xcolor}
\usepackage{lipsum}    
\usepackage{subfig}    
\usepackage[T1]{fontenc}
\usepackage{mathrsfs}
\usepackage{algorithm}

\newcommand{\eref}[1]{Eq.~\ref{#1}}
\newcommand{\fref}[1]{Figure~\ref{#1}}
\newcommand{\tref}[1]{Table~\ref{#1}}
\newcommand{\sref}[1]{Section~\ref{#1}}
\newcommand{\cref}[1]{Chapter~\ref{#1}}
\newtheorem{definition}{Definition}

\newtheorem{fact}{Fact}

\newcommand{\ben}{\begin{enumerate}}
\newcommand{\een}{\end{enumerate}}

\newcommand{\bo}{\textbf}

\newcommand{\x}{\textbf{x}}
\newcommand{\X}{\textbf{X}}

\newcommand{\A}{\textbf{A}}

\newcommand{\bP}{\textbf{P}}
\newcommand{\bD}{\textbf{D}}

\newcommand{\M}{\textbf{M}}

\newcommand{\cC}{\mathscr{C}}
\newcommand{\e}{\textbf{e}}

\newcommand{\kmeans}{\textit{k}-means\,}
\DeclareMathOperator*{\argmin}{arg\,min}

\begin{document}

\title{A Flexible Iterative Framework for Consensus Clustering 
}
\author{
Carl Meyer,
Shaina Race
}
\maketitle

\begin{abstract}
A novel framework for consensus clustering is presented which has the ability to determine both the number of clusters and a final solution using multiple algorithms. A consensus similarity matrix is formed from an ensemble using multiple algorithms and several values for $k$.   A variety of dimension reduction techniques and clustering algorithms are considered for analysis.  For noisy or high-dimensional data, an iterative technique is presented to refine this consensus matrix in way that encourages algorithms to agree upon a common solution.  We utilize the theory of nearly uncoupled Markov chains to determine the number, $k$ , of clusters in a dataset by considering a random walk on the graph defined by the consensus matrix. The eigenvalues of the associated transition probability matrix are used to determine the number of clusters. This method succeeds at determining the number of clusters in many datasets where previous methods fail. On every considered dataset, our consensus method provides a final result with accuracy well above the average of the individual algorithms.  
\end{abstract}

\section{Introduction}
Cluster analysis is an important tool used in hundreds of data mining applications like image segmentation, text-mining, genomics, and biological taxonomy. Clustering allows users to find and explore patterns and structure in data without prior knowledge or training information. Hundreds (if not thousands) of algorithms exist for this task but no single algorithm is guaranteed to work best on any given class of real-world data. This inconsistency of performance in cluster analysis is not unique to the clustering algorithms themselves. In fact, the dimension reduction techniques that are expected to aid these algorithms by revealing the cluster tendencies of data also tend to compete unpredictably, and it is difficult to know beforehand which low-dimensional approximation might provide the best separation between clusters.  Having many tools and few ways to make an informed decision on which tool to use, high-dimensional cluster analysis is doomed to become an ad hoc science where analysts blindly reach for a tool and hope for the best. Cluster analysis is not the first type of data mining to encounter this problem. Data scientists were quick to develop ensemble techniques to escape the unreliability of individual algorithms for tasks like prediction and classification. Ensemble methods have become an integral part of many areas of data mining, but for cluster analysis such methods have been largely ignored. 

An additional problem stems from the fact that the vast majority of these algorithms require the user to specify the number of clusters for the algorithm to create. In an applied setting, it is unlikely that the user will know this information before hand. In fact, the number of distinct groups in the data may be the very question that an analyst is attempting to answer. Determining the number of clusters in data has long been considered one of the more difficult aspects of cluster analysis. This fact boils down to basics: what is a cluster? How do we define what should and should not count as two separate clusters? Our approach provides an original answer this question: A group of points should be considered a cluster when a variety of algorithms \textit{agree} that they should be considered a cluster. If a majority of algorithms can more or less agree on how to break a dataset into two clusters, but cannot agree on how to partition the data into more than two clusters, then we determine the data has two clusters. This is the essence of the framework suggested herein.  

Our purpose is to address both problems: determining the number of clusters and determining a final solution from multiple algorithms. We propose a consensus method in which a number of algorithms form a voting ensemble, proceeding through several rounds of elections until a majority rule is determined.  This allows the user to implement many tools at once, increasing his or her confidence in the final solution. 

\section{Consensus Clustering}
\subsection{Previous Proposals for Consensus Clustering}
\label{oldconsensus}
In recent years, the consensus idea has been promoted by many researchers \cite{consensusberman,consensusgene,consensusspectral,chuck, consensusnam, consensuszeng, consensusLS, consensusjain, consensusfern,consensusfilkov,consensusstrehl,amaral, mirkin2}. The main challenge to ensemble methods using multiple algorithms is generally identified to be the wide variety in the results produced by different algorithms due to the different cluster criteria inherent in each algorithm. Thus any direct combination of results from an ensemble will not often generate a meaningful result \cite{consensusqian,consensuskot}. 

Most often the consensus problem has been formulated as an optimization problem, where the optimal clustering, $\cC^*$, minimizes some relative distance metric between $\cC^*$ and all of the clusterings $\cC_i$ in the ensemble. There are many ways to define the distance between two clusterings, for example one could take the minimum number of elements that need to be deleted for the two partitions to become identical \cite{gusfield}. Using $d(\cC_1,\cC_2)$ to denote some measure of distance between two different clusterings, we'd write
\begin{equation}
\cC^*=\argmin_{\cC} \sum_{i=1}^N d(\cC_i,\cC).
\label{medianobj}
\end{equation}
 This problem is known as the \textit{median partition problem} in the literature and dates back to the 1965-`74  work of R{\'e}gnier (\cite{regnier}) and Mirkin (\cite{mirkin}) \cite{consensusfilkov}. Alternatively, some authors use a relative validity metric like the normalized mutual information $NMI(\cC_i,\cC)$ in place of a distance function and attempt to maximize the objective function in \eref{medianobj} \cite{consensusstrehl}. The median partition problem was shown by Krivanek and Moravek, and also by Wakabayashi, to be NP-complete \cite{consensusfilkov}, but many heuristics have since been proposed to find approximate solutions \cite{consensusstrehl, consensusfilkov, consensusberman, consensusLS}.
 
 We believe that these methods are bound to suffer because each clustering in the ensemble is given equal importance. Suppose we had 4 perfect clusterings and 1 terribly inaccurate clustering. These methods would not take into account the fact that the majority of the algorithms share 100\% agreement on a perfect clustering, and instead may shift the optimal clustering away from perfection towards inaccuracy. Thus, we feel that the optimization in \eref{medianobj} leads to a ``middle-of-the-road'' solution or a \textit{compromise} between algorithms, rather than a solution of ``agreement'' or consensus.  In our method, the clustering algorithms act as a voting ensemble and continually move through a series of elections until a desired level of consensus is reached. Additionally, we introduce a parameter of intolerance, which allows the user to impose a level of agreement that must be reached between algorithms in order to accept a cluster relationship between objects. 

\subsection{The Consensus Matrix}
\label{consensusmatrix}
\begin{figure}[h!]
\centering
\includegraphics[width=0.7\linewidth]{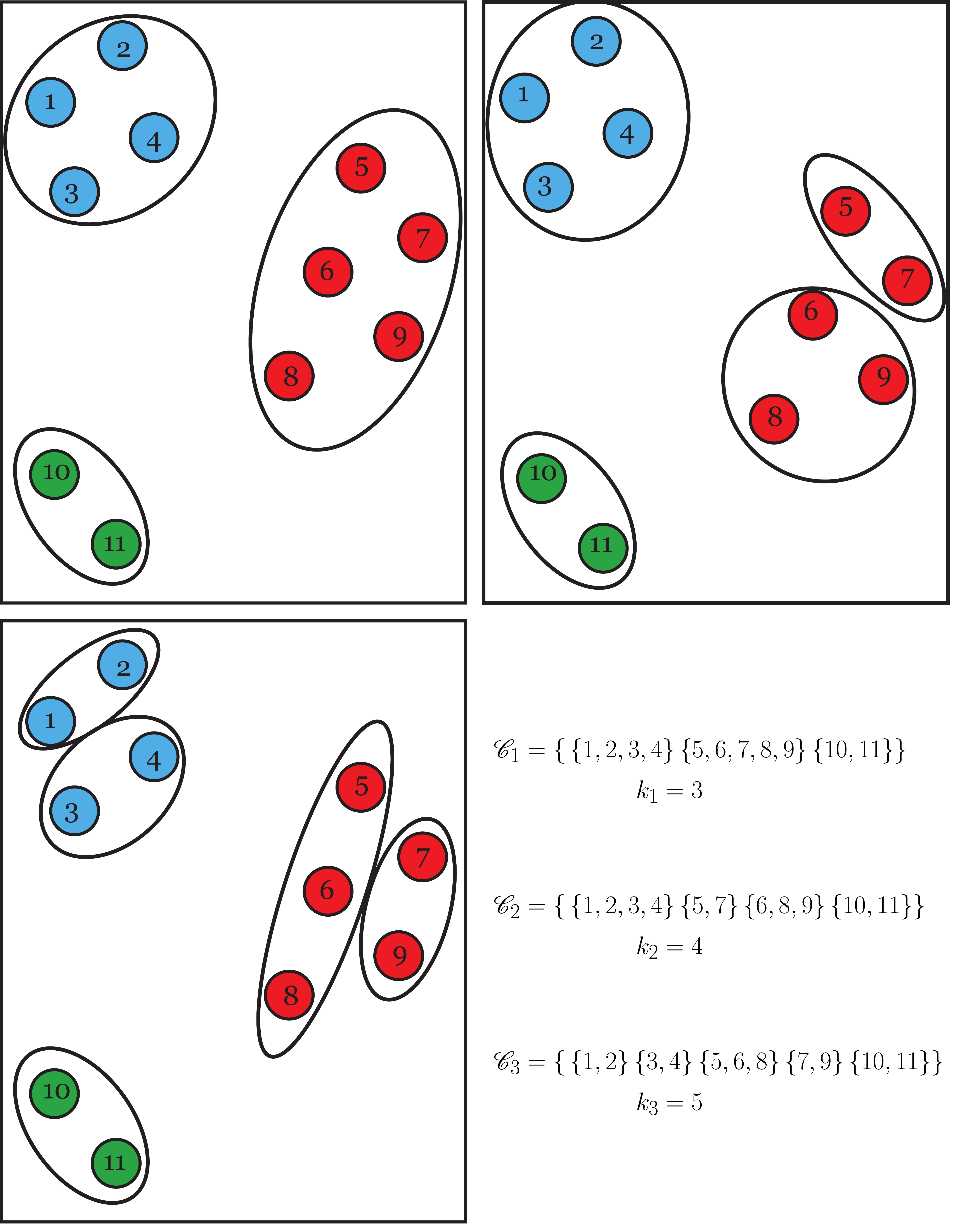}
\caption{Example of an Ensemble of $N=3$ Clusterings}
\label{ensembleex}
\end{figure}

  To begin, we introduce some notation. Since consensus methods combine multiple solutions from multiple algorithms (or multiple runs of the same algorithm), we start with a \textbf{cluster ensemble}. A cluster ensemble, $\cC=\{\cC_1,\cC_2,\dots,\cC_N\}$, is a set of $N$ clusterings of the $n$ data objects $\x_1,\x_2,\dots,\x_n$. That is, each clustering $\cC_j$ in the ensemble is a $k_j$-way partition of the data, composed of individual clusters,
$$\cC_j = [C_1,C_2, \dots, C_{k_j}],$$
 where the number of clusters $k_j$ in each clustering may be allowed to vary. In \fref{ensembleex}, we illustrate a simple example with $N=3$ clusterings.
 
The information from a cluster ensemble is then recorded in a \textbf{consensus matrix}.
\begin{definition}[The Consensus Matrix]
Given a cluster ensemble, $$\cC =\{\cC_1,\cC_2,\dots,\cC_N\}, $$ of $n$ data points $\x_1,\x_2,\dots,\x_n$, the \textbf{consensus matrix} $\bo{M}$ is an $n\times n$ matrix such that
$$\M(\cC)_{ij} = \# \mbox{ of times object } \x_i \mbox{ was placed in the same cluster as } \x_j \mbox{ in the ensemble }\cC.$$
\label{cmatrix}
\end{definition} 

One might prefer to think of the consensus matrix as the sum of individual adjacency matrices for each clustering in the ensemble. For a given clustering $\cC_i$ we could define an adjacency matrix, $\A_i$ as
$$\A_{ij} = \left\{ \begin{array}{ll}
1 &\mbox{ if object } \x_i \mbox{ was clustered with } \x_j \\
0 &\mbox{ otherwise }
\end{array}
\right.
$$
Then the consensus matrix $\M$ would be the sum of the adjacency matrices of each clustering in the ensemble:
$$\M(\cC) = \sum_{i=1}^N \A_i. $$\\

\begin{figure}[h!]
\centering
\includegraphics[width=0.5\linewidth]{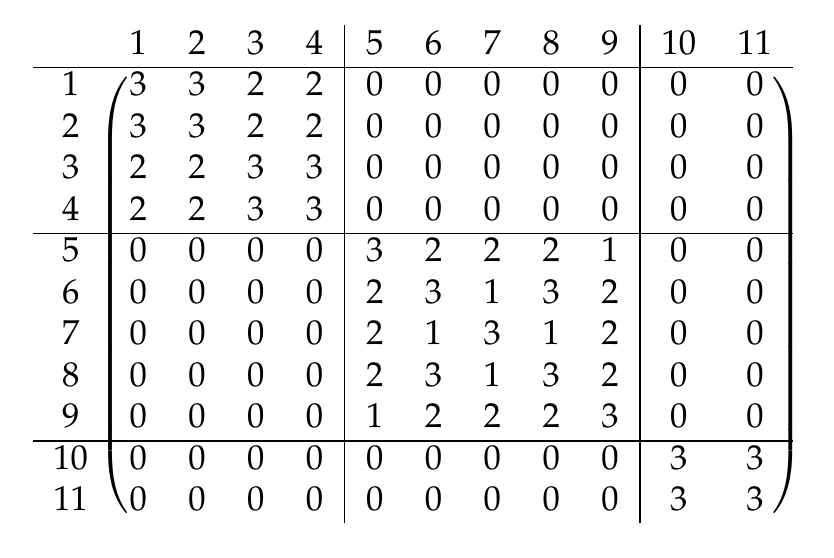}
\caption{The Consensus Matrix for the Ensemble in \fref{ensembleex}}
\label{consensusex}
\end{figure}

As an example, the consensus matrix for the ensemble depicted in \fref{ensembleex} is given in \fref{consensusex}.

The consensus matrix from \fref{consensusex} is very interesting because the ensemble that was used to create it had clusterings for various values of $k$. The most reasonable number of clusters for the colored circles in \fref{ensembleex} is $k^*=3$. The 3 clusterings in the ensemble depict $k_1=3, k_2=4, \mbox{ and } k_3=5$ clusters. However, the resulting consensus matrix is clearly block-diagonal with $k^*=3$ diagonal blocks! This is not an isolated phenomenon, in fact it is something we should expect from our consensus matrices if we labor under the following reasonable assumptions:
 \begin{itemize}
\item If there are truly $k$ distinct clusters in a given dataset, and a clustering algorithm is set to find $\tilde{k}>k$ clusters, then the $k$ ``true'' clusters will be broken apart into smaller clusters to make $\tilde{k}$ total clusters.
\item Further, if there is no clear ``subcluster" structure, meaning the original $k$ clusters do not further break down into meaningful components, then different algorithms will break the clusters apart in different ways. 
\end{itemize}  

This block-diagonal structure is the subject of \sref{perron}.


\subsubsection{Benefits of the Consensus Matrix}
\label{benefits}
\begin{figure}[h!]
\centering
\begin{subfigure}{.49\textwidth}
\centering
\includegraphics[width=0.97\linewidth]{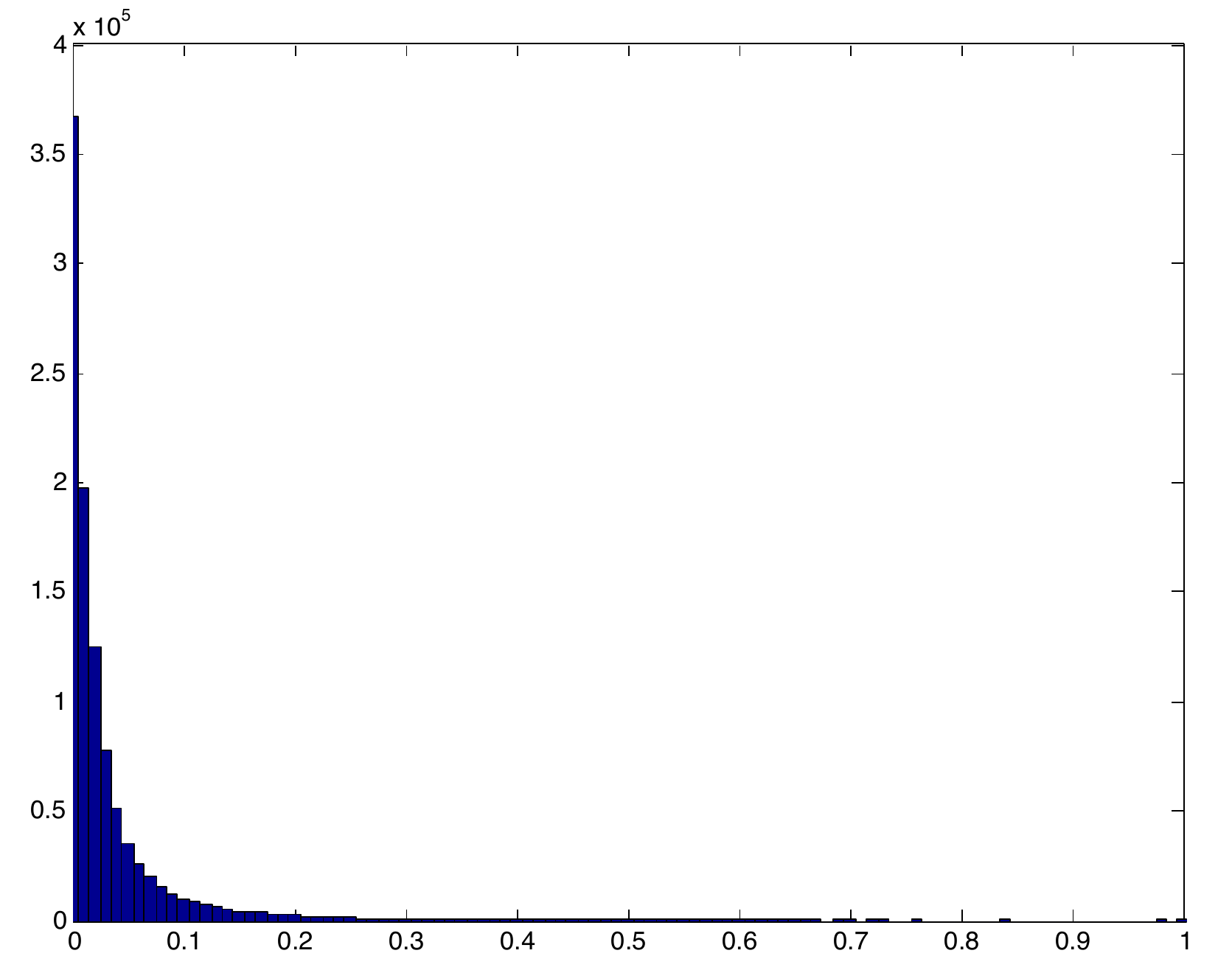}
\caption{Histogram of 1 million random entries in a cosine similarity matrix}
\label{MCCcoshist}
\end{subfigure}
\begin{subfigure}{.49\textwidth}
\centering
\includegraphics[width=0.97\linewidth]{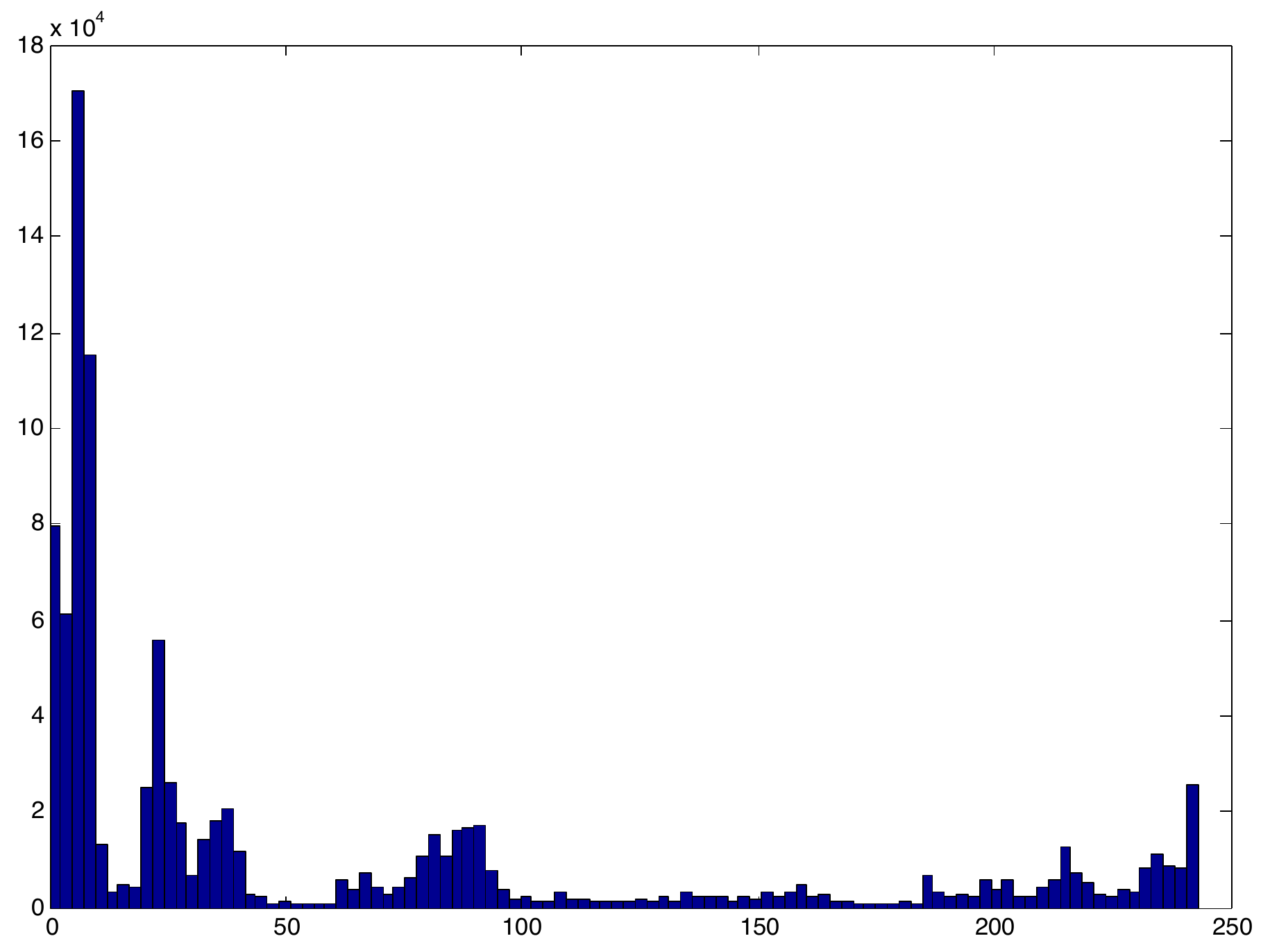}
\caption{Histogram of same 1 million entries in consensus similarity matrix}
\label{MCCconshist}
\end{subfigure}
\caption{Distribution of Similarity Values in Cosine vs. Consensus Matrix}
\label{matrixdist}
\end{figure}

As a similarity matrix, the consensus matrix offers some benefits overs traditional approaches like the Gaussian or Cosine similarity matrices. One problem with these traditional methods is the curse of dimensionality: In high dimensional spaces, distance and similarity metrics tend to lose their meaning. The range of values for the pairwise distances tightens as the dimensionality of the space grows, and little has been done to address this fact. In \fref{matrixdist} we show the distribution of similarity values for the same 1 million entries in a consensus matrix compared to the cosine similarity matrix. As you can see, the consensus approach allows a user to witness some very high levels of similarity in high-dimension data, whereas the cosine similarities have a much smaller range. The dataset, which is more formally introduced in \sref{mcc}, is the Medlars-Cranfield-CISI document collection ($\approx 4,000$ documents) \cite{kogan}. Such contrast is typical among high-dimensional datasets.

An additional benefit is that entries in the consensus matrix have depth. By this, we mean that they result from summing entries in adjacency matrices output by individual clustering algorithms, so more information is available about the \textit{meaning} of each similarity value. The cosine of the angle between two data vectors $\x_i$ and $\x_j$ may tell us something about their correlation, but knowing, for instance that these two objects were clustered together by all algorithms with $\tilde{k} \leq 5$, by some algorithms with $6\leq\tilde{k}\leq7$, and never when $\tilde{k}\geq 7$, provides a depth of insight not previously considered. While we do not use this information explicitly in our analysis, it may be beneficial in practical research.

The greatest benefit of using a consensus matrix for clustering is that it provides superior information about clustering within the data. This has been demonstrated time and again in the literature \cite{chuck, phdthesis,siamicc,consensusfred, consensusgene,consensusjain, consensusspectral, consensusstrehl,jain50, mirkin2}. We add to the pile of evidence for this statement with the experiments in this paper.

\section{Iterative Consensus Clustering (ICC)}
\label{icc}
The consensus approach outlined herein is based on the work in \cite{consensusspectral,consensusjain,thesis,chuck,consensusfred} where the consensus matrix is treated as similarity matrix and used as input to a clustering algorithm to reach a final solution. In \cite{consensusspectral} the authors suggest using many runs of the \kmeans algorithm, initialized randomly, to build the consensus matrix and then using a spectral clustering method, such as normalized cut (NCut) \cite{shi}, to determine a final solution. In \cite{consensusjain, jain50,consensusfred}, the approach is again to build a consensus matrix using many runs of the \kmeans algorithm and then to cluster the consensus matrix with one final run of \kmeans. In \cite{chuck} a consensus matrix is formed via \kmeans and then used as input to the stochastic clustering algorithm (SCA). While all these methods provide better results than individual algorithms, they still rely on a single algorithm to make both the consensus matrix and the final decision on cluster membership. 

Our method uses a variety of algorithms, rather than just \kmeans, to create the initial cluster ensemble. In addition, each algorithm is paired with different dimension reductions because it is often unclear which dimension reduction gives the best configuration of the data; each lower dimensional representation has the potential to separate different sets of clusters in the data. In this way, we essentially gather an initial round of \textit{votes} for whether or not each pair of objects $(\x_i,\x_j)$ belong in the same cluster. These votes are collected in a consensus matrix $\M$ as defined in Definition \ref{cmatrix}.

\subsection{Clustering Algorithms Used for Ensembles}
\label{algs}
The experiments contained herein use a number of different clustering algorithms. The details of these popular algorithms are omitted due to space considerations but we refer the reader to the following resources:
\paragraph{Data Clustering Algorithms}
\begin{enumerate}
\item Spherical $k$-means (solution with lowest objective value from 100 runs, randomly initialized) \cite{datamining}
\item PDDP: Principal Direction Divisive Partitioning \cite{pddp} 
\item PDDP-$k$-means:  $k$-means with initial centroids provided by PDDP clusters \cite{phdthesis}
\item NMFcluster: $k$ dimensional Nonnegative Matrix Factorization for Clustering \cite{NMFcluster} 
\end{enumerate}
\paragraph{Graph Clustering Algorithms (Requiring a similarity matrix) }
\begin{enumerate}
\item PIC: Power Iteration Clustering \cite{poweriteration}
\item NCUT: Normalized Cuts according to Shi and Malik \cite{shi}
\item NJW: Normalized Cuts according to Ng, Jordan, and Weiss \cite{ng}
\end{enumerate}

 Each algorithm in our ensemble is assumed to be \textit{reasonable}, making good choices on cluster membership most of the time. It is inevitable that each of the clustering algorithms will make mistakes, particularly on noisy data, but it is assumed that rarely will the majority of algorithms make the \textit{same} mistake. To account for this error, we introduce an \textit{intolerance parameter}, $\tau$, for which entries in the consensus matrix $\M_{ij} < \tau N$ will be set to zero. In other words, $\tau$ is the minimum proportion of algorithms that must agree upon a single cluster relationship $(\x_i,\x_j)$ in order to keep those ``votes'' in the consensus matrix. 

\begin{figure}[h!]
\centering
\includegraphics[width=0.7\linewidth]{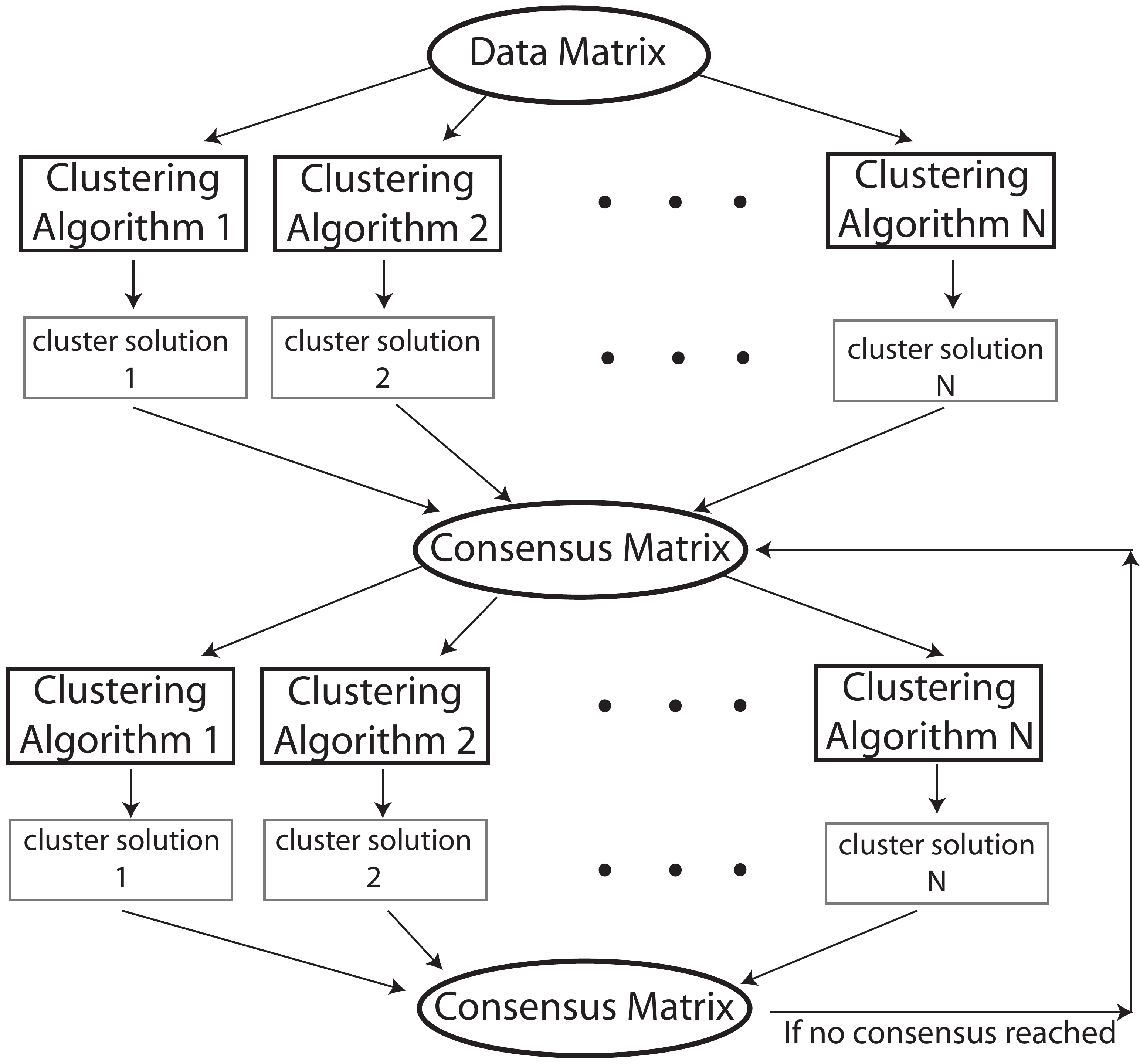}
\caption{Iterated Consensus Clustering (ICC) Process}
\label{consensusflowchart}
\end{figure}

After the initial consensus matrix is formed, we use it as input to each of the clustering algorithms again. Essentially we start a debate between algorithms, asking each of them to use the collective votes of the ensemble to determine a second solution. Again these solutions are collected in a consensus matrix and the process repeats until a simple majority of algorithms agree upon one solution. Once the majority of algorithms have chosen a common solution, we say the algorithms have \textit{reached consensus} and call that resulting solution the \textbf{final consensus solution}. This process is illustrated in the flow chart in \fref{consensusflowchart}.

\subsection{Example with Medlars-Cranfield-CISI Text Collection}
\label{mcc}
 To illustrate the effectiveness of this procedure, we consider a combination of 3 text datasets used frequently in the information retrieval literature. For simplicity, we assume the number of clusters is known a priori. In \sref{perron} this information will be extracted from the data.  The combined Medlars-Cranfield-CISI (MCC) collection consists of nearly 4,000 scientific abstracts from 3 different disciplines. These 3 disciplines (Medlars = medicine, Cranfield = aerodynamics, CISI = information science) form 3 natural clusters in the data  \cite{surveytextmining,kogan}.  
 
 The document data is high-dimensional with $m \approx 11,000$ features (words). As a result, clustering algorithms tend to run slowly on the raw data. Thus, we reduce the dimensions of the data using 3 preferred algorithms:
\begin{itemize}
\item[1.] Nonnegative Matrix Factorization (NMF) by Alternating Constrained Least Squares (ACLS) \cite{AppliNMF}
\item[2.] Singular Value Decomposition (SVD) \cite{drineassvd, LSI}
\item[3.] Principal Components Analysis (PCA) \cite{PCA}
\end{itemize}

One should realize that PCA is, in fact, a Singular Value Decomposition of Data under z-score normalization. However, in practice, these two decompositions generally provide quite different results, particularly for high-dimensional data. 

For each dimension reduction technique, the number of features is reduced from $m = 11,000$ to $r=5, 10,\mbox{ and } 20$ creating a total of 9 input data sets. On each input dataset, 3 different clustering methods were used to cluster the data:
\begin{itemize}
\item[1.] $k$-means
\item[2.] PDDP
\item[3.] PDDP-$k$-means
\end{itemize}

The accuracy (proportion of documents classified correctly \cite{phdthesis}) of each algorithm on each data input are given in \tref{mccresults}.

\begin{table}[h!]
\begin{tabular}{|c | l|  c| c| c|}
\hline
 \# Features & Algorithm & NMF input & SVD input & PCA input \\
 \hline 
 				& \kmeans			&0.8741&  0.7962&0.8260\\
 ($r=5$) 	& PDDP					& 0.9599&  0.9049& 0.9026\\
 				& PDDP-\kmeans &0.9599& 0.9049& 0.9026\\
 \hline
 				& \kmeans 			& 0.8628&0.8268&0.8286\\
 ($r=10$) 	& PDDP				& 0.9764& 0.9774& 0.9481\\
 				& PDDP-\kmeans &0.9764& 0.9774&0.9481\\
 \hline 
 				& \kmeans			& 0.8530&0.8263&0.8281\\
 ($r=20$) 	& PDDP				& 0.9722&0.9802&0.9478\\
 				& PDDP-\kmeans &0.6114&0.9802&0.9478\\
 \hline 
 \multicolumn{5}{c}{Average Accuracy of All Clusterings: 0.90} \\
 \hline

\end{tabular}
\caption{Accuracy Results for 3 Clustering Algorithms on 9 Low Dimensional Representations of the Medlars-Cranfield-CISI text data}
\label{mccresults}
\end{table}
The accuracy of the results ranges from 61\% ($>$1,500 misclassified) to 98\% (78 misclassified). A reasonable question one might ask is this: Why not choose the solution with the lowest \kmeans objective value? The biggest problem with this boils down to the curse of dimensionality: the distance measures used to compute such metrics lose their meaning in high-dimensional space \cite{phdthesis, mirkin2}. The only comparison we could make between clusterings would be with the full dimensional data and, surprisingly, the objective function values for the minimum accuracy solution is approximately equal to the maximum accuracy solution! Other internal metrics, like the popular Silhouette coefficient \cite{datamining} suffer from the same problem. One must be very careful when attempting to compare high-dimensional clusterings with such metrics.

Our suggestion is instead to compile the clusterings from \tref{mccresults} into a consensus matrix, cluster that consensus matrix with multiple algorithms, and repeat that process until the majority of the algorithms agree upon a solution. This can be done with or without dimension reduction on the consensus matrix. For simplicity, we'll proceed without reducing the dimensions of the consensus matrix, but we will include an additional clustering algorithm, NMFCluster, which was not well suited for the analysis on the low-dimensional representations in \tref{mccresults}. \tref{mccconsensus} provides the accuracy of these 4 clustering algorithms on the consensus matrices through iteration. Boxes are drawn around values to indicate a common solution chosen by algorithms. A final consensus solution is found in the third iteration with 3 of the 4 algorithms agreeing upon a single solution. The accuracy of this final consensus solution is much greater than the average of all the initial clustering results in \tref{mccresults}. Such a result is typical across all datasets considered.

\begin{table}[h!]
\centering
\begin{tabular}{|c| c | c| c| }
\hline
  Algorithm & Consensus Iter 1 &  Consensus Iter 2 & Consensus Iter 3  \\
 \hline
 PDDP& 0.939 &  0.969 &  0.969 \\
  PDDP-\kmeans & 0.954  & \framebox{0.966} & \framebox{0.966} \\
 NMFcluster & 0.969&0.954 & \framebox{0.966} \\
 \kmeans & 0.966 &\framebox{0.966} & \framebox{0.966}  \\
   \hline
\end{tabular}
\caption{Medlars-Cranfield-CISI text collection: Accuracies for 4 Clustering Algorithms on Consensus Matrices through Iteration} 
\label{mccconsensus}
\end{table}

\subsection{Perron Cluster Analysis}
\label{perron}
In \sref{consensusmatrix} an example was given that alluded to our methodology for determining the number of clusters. We approach this task using Perron-cluster methodology \cite{chuck,fischer,perroncluster,siamicc,phdthesis} on the consensus similarity matrix. Perron-cluster analysis involves the examination of eigenvalues of a \textit{nearly uncoupled} or \textit{nearly completely reducible} Markov chain. We consider the transition probability matrix $\bP$ of a random walker on the graph defined by the consensus matrix $\M$:
$$\bP=\bD^{-1}\M$$ where $\bD$ is a diagonal matrix containing the row sums of $\M$: $\bD=diag(\M \e).$ 
According to our assumptions in \sref{cmatrix}, there exists some simultaneous permutation of rows and columns of our consensus matrix such that the result is \textit{block-diagonally dominant}. By this we essentially mean that $\bP$ (after row and column permutation) is a perturbation of a block-diagonal matrix $\bo{B}$, such that

\begin{equation}
\bP=\bo{B}+\bo{E} = \left[ 
\begin{array}{ccccc}
\bo{B}_{11} & \bo{E}_{12} & \bo{E}_{13}& \dots  & \bo{E}_{1k} \\
\bo{E}_{21}   & \bo{B}_{22} & \bo{E}_{23} & \dots & \bo{E}_{2k} \\
\bo{E}_{31}   & \bo{E}_{32} & \bo{B}_{33} & \ddots & \bo{E}_{3k} \\
\vdots& \vdots& \vdots & \ddots & \vdots  \\
\bo{E}_{k1} & \bo{E}_{k2}& \bo{E}_{k3} & \dots & \bo{B}_{kk}
\end{array}
\right]
 \label{bdd}
 \end{equation}
 where the off-diagonal blocks, $\bo{E}_{ij}$, are much smaller in magnitude than the the diagonal blocks. In fact, the entries in the off-diagonal blocks are small enough that the diagonal blocks are \textit{nearly stochastic}, i.e. $\bo{B}_{ii} \e \approx 1$ for $i=1,2,\dots,k$.  A transition probability matrix taking this form describes a \textbf{nearly uncoupled} or \textbf{nearly completely reducible} Markov Chain.
 
 The degree to which a matrix is considered nearly uncoupled is dependent on one's criteria for measuring the level of \textit{coupling} (interconnection) between the \textit{aggregates} (clusters of states) of the Markov chain \cite{fischer,meyernumc,chuckthesis}. In \cite{meyernumc}, the \textit{deviation from complete reducibility} is defined as follows:
  
 \begin{definition}[Deviation from Complete Reducibility]
 For an $m\times n$ irreducible stochastic matrix with a $k$-cluster partition
 $$\bP = \left[ 
\begin{array}{ccccc}
\bo{P}_{11} & \bo{P}_{12} & \bo{P}_{13}& \dots  & \bo{P}_{1k} \\
\bo{P}_{21}   & \bo{P}_{22} & \bo{P}_{23} & \dots & \bo{P}_{2k} \\
\bo{P}_{31}   & \bo{P}_{32} & \bo{P}_{33} & \ddots & \bo{P}_{3k} \\
\vdots& \vdots& \vdots & \ddots & \vdots  \\
\bo{P}_{k1} & \bo{P}_{k2}& \bo{P}_{k3} & \dots & \bo{P}_{kk}
\end{array}
\right]$$
 the number $$\delta=2\max_{i} \|\bP_{i*}\|_{\infty},$$ where $\bP_{i*}$ represents a row of blocks, is called the \textbf{deviation from complete reducibility.}
 \end{definition}
  
It is important to point out that the parameter $\delta$, or any other parameter that measures the level of coupling between clusters in a graph (like those suggested in \cite{fischer,chuckthesis}) cannot be computed without prior knowledge of the clusters in the graph. Such parameters are merely tools for the perturbation analysis, used to present the following important fact regarding the spectrum of block-diagonally dominant stochastic matrices \cite{fischer,kato,chuck,meyernumc,perroncluster, stewartnumc}:\\ 
 
 \begin{fact}[The Spectrum of a Block-Diagonally Dominant Stochastic Matrix \cite{fischer,meyernumc,perroncluster,chuck} ]
For sufficiently small $\delta \neq 0$, the eigenvalues of $\bP(\delta)$ are continuous in $\delta$, and can be divided into 3 parts:
\begin{itemize}
\item[1.] The Perron root, $\lambda_1(\delta)=1$,
\item[2.] a cluster of $k-1$ eigenvalues $\lambda_2(\delta),\lambda_3(\delta),\dots,\lambda_k(\delta)$ that approach 1 as $\delta \to 0$ (known as the \textbf{Perron cluster}), and
\item[3.] the remaining eigenvalues, which are bounded away from 1 as $\delta \to 0$.
\end{itemize}
\end{fact}
 
 In order to recover the number of blocks (clusters), we simply examine the eigenvalues of the stochastic matrix $\bP=\bD^{-1}\M$ and count the number of eigenvalues in the Perron cluster, which is separated from the remaining eigenvalues by the \textbf{Perron gap}, the largest difference between consecutive eigenvalues $\lambda_k$ and $\lambda_{k+1}$. The size of this gap is determined by the level of uncoupling in the graph, with larger gaps indicating more nearly uncoupled structures \cite{chuck}.

\subsection{Perron Cluster Analysis for Consensus Matrices }
\label{perronconsensus}
 To build the consensus similarity matrix, we use one or more algorithms to cluster the data into a varying number of clusters.  We set the algorithm(s) in our ensemble to find $\tilde{k}_1,\tilde{k}_2, \dots, \tilde{k}_J$ clusters in the data.  The choice of the values $\tilde{k}_i$ is for the user, but we suggest choosing these values such that they might over-estimate the number of clusters but remain less than $\sqrt{n}$.  We then construct the consensus similarity matrix $\M$ from the resulting clusterings, examine the eigenvalues of the transition probability matrix $\bP=\bD^{-1}\M$, and count the number of eigenvalues near $\lambda_1 = 1$ by locating the largest gap in the eigenvalues.

\begin{figure}[h!]
\centering
\includegraphics[width=0.5\linewidth]{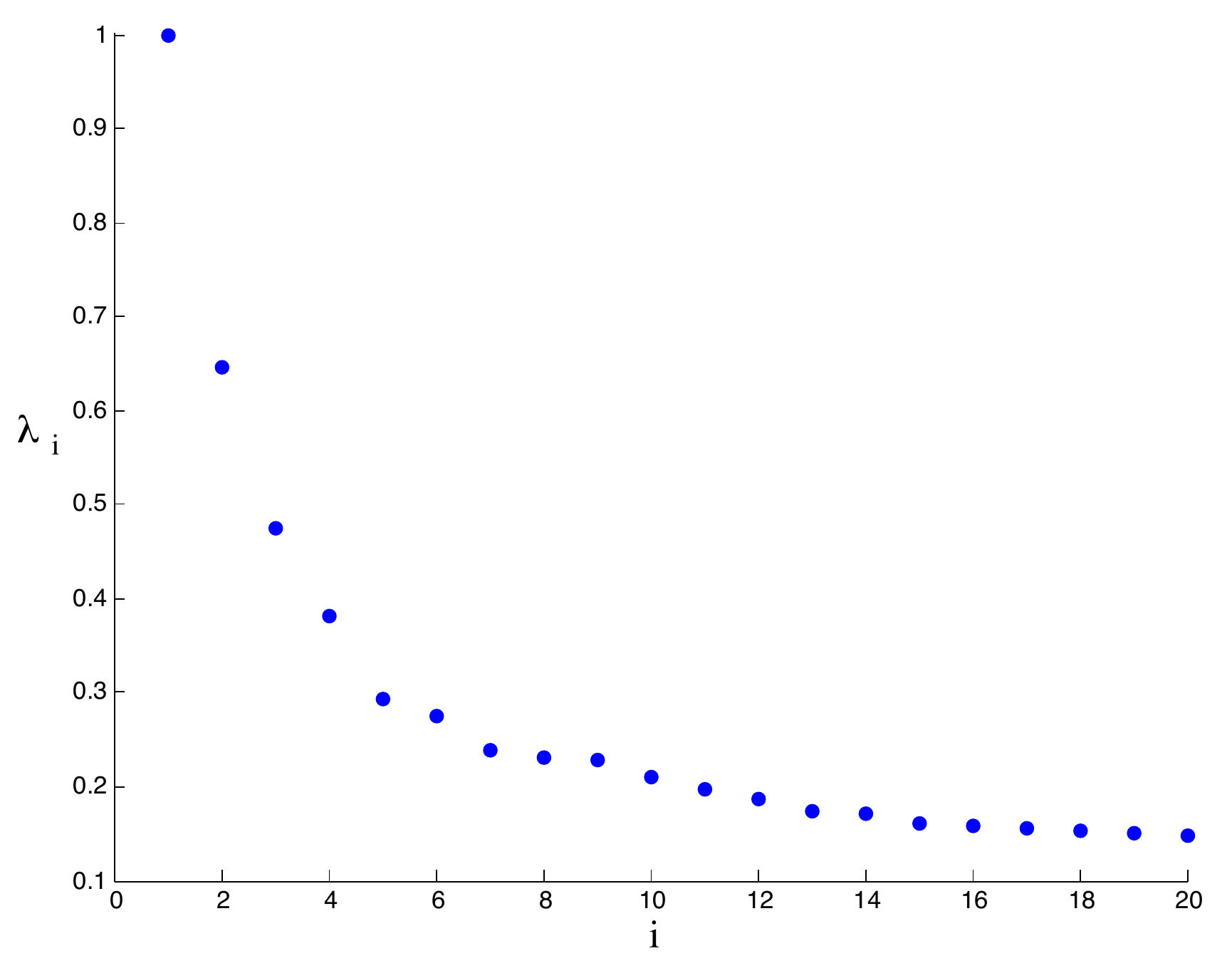}
\caption{Dataset MCC: 20 Largest Eigenvalues Found Using Cosine Similarity Matrix}
\label{MCCcosineEigs}
\end{figure}

It is sometimes helpful in this phase of the algorithm to consider the use of an intolerance parameter $\tau$ in the construction of the consensus matrix because we may have allowed the algorithms to partition the data into fewer clusters than actually exist in the data. We return to the Medlars-Cranfield-CISI (MCC) collection discussed in \sref{mcc} as an example where traditional SSE plots generally fail to provide a clear picture of how many clusters may be optimal \cite{phdthesis}.  Before discussing the results of our method, we first look at the eigenvalues of the transition probability matrix that would result from using cosine as a measure of similarity (this is the most common similarity matrix used for text collections in the spectral clustering literature).  The largest eigenvalues of this $3891 \times 3891$ matrix are displayed in \fref{MCCcosineEigs}. The plot shows only one eigenvalue in the Perron cluster and therefore, as with the other methods discussed in \cite{phdthesis}, no information is gathered about the number of clusters in the data.

\begin{figure}[h!]

\centering
\begin{subfigure}{.47\textwidth}
\centering
\includegraphics[width=0.9\linewidth]{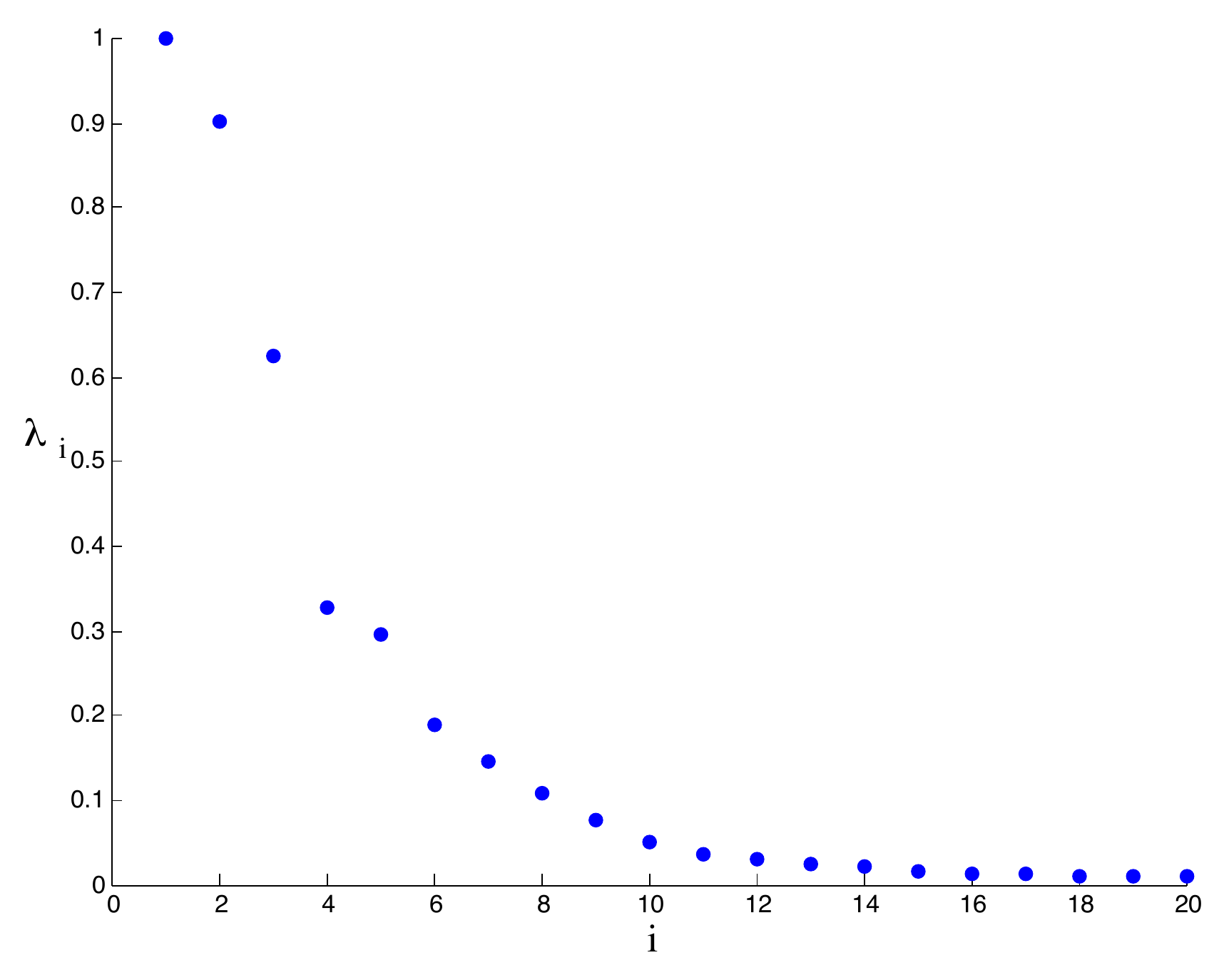}
\caption{No Intolerance ($\tau = 0.0$)}
\label{MCCnodropEigs}
\end{subfigure}
\begin{subfigure}{.47\textwidth}
\centering
\includegraphics[width=0.9\linewidth]{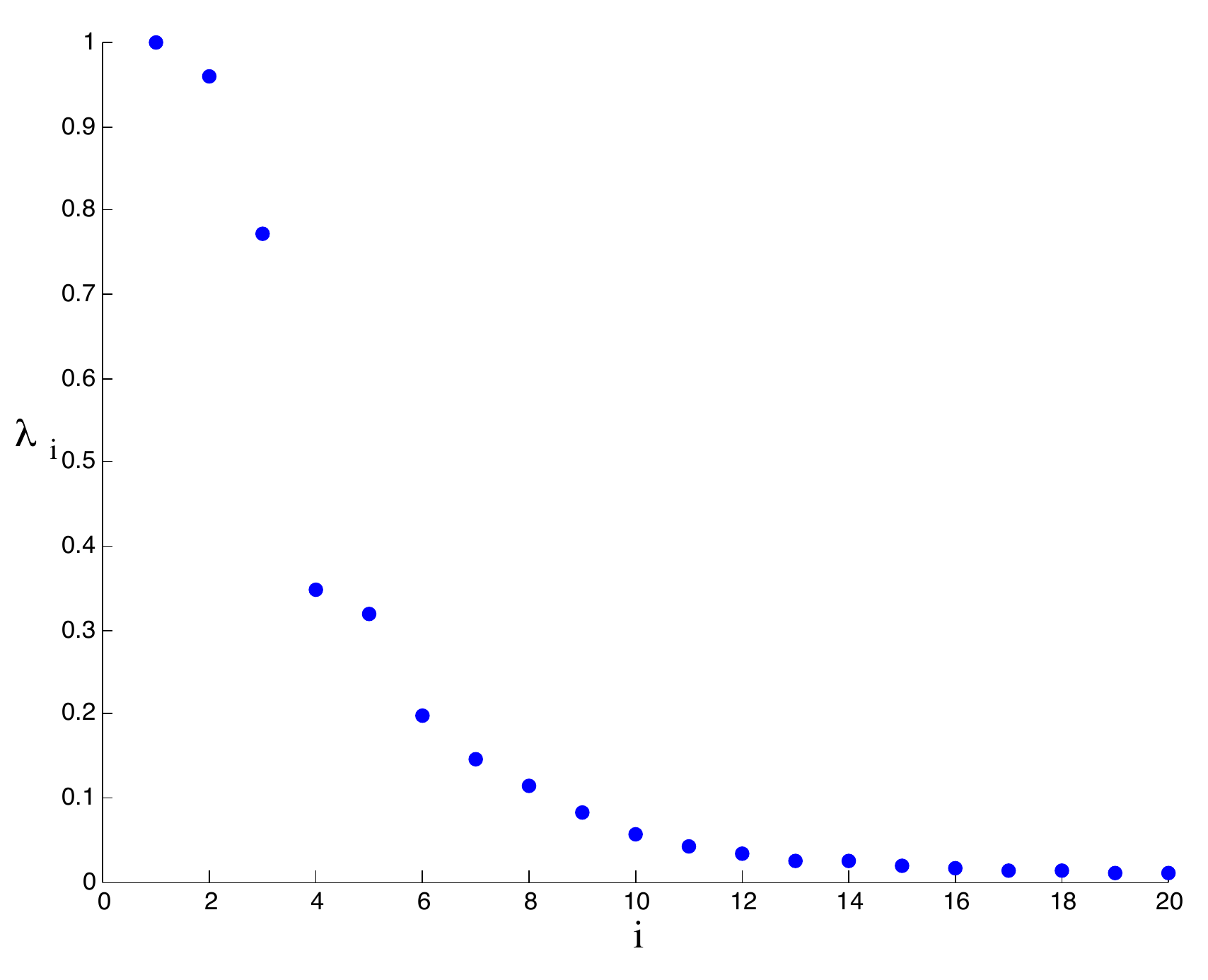}
\caption{10\% Intolerance ($\tau = 0.1$)}
\label{MCCdrop10Eigs}
\end{subfigure}

\caption{Dataset MCC: 20 largest eigenvalues found using consensus similarity matrices with (right) and without (left) the intolerance parameter $\tau$. Ensemble of 30 algorithms, each clustering data into $\tilde{k}=2,3,\dots,10$ clusters}
\label{mccicceigs}
\end{figure}

Now we look at the eigenvalues of the transition probability matrix associated with a consensus similarity matrix. This consensus matrix was built from an ensemble of various algorithms paired with different dimension reductions and different levels of dimension reduction. All 3 of the authors' preferred dimension reduction techniques (NMF, PCA, SVD) were used to reduce the dimensions of the data to $r=5,10, \mbox{ and } 20$ dimensions, creating a total of 10 data inputs (including the raw high-dimensional data) for each clustering algorithm. Three different clustering methods were used to cluster each data input: PDDP, spherical \kmeans initialized randomly, and spherical \kmeans initialized with centroids from the clusters found by PDDP. Counting every combination of dimension reduction and clustering procedure, the ensemble had 30 algorithms at work. For each of the 30 algorithms, $\tilde{k}=2,3,4,\dots,10$ clusters were determined and the resulting 270 clusterings were collected in the consensus matrix $\M$. We show in \fref{mccicceigs} side-by-side images showing the eigenvalues of the transition probability matrix associated with the consensus similarity matrix \textit{with} and \textit{without} use of the intolerance parameter $\tau$. Particularly with text and other high-dimensional data, this intolerance parameter, by removing extraneous connections in the consensus graph, encourages a \textit{nearly uncoupled} structure in the clustering results \cite{siamicc,phdthesis}. This \textit{uncoupling effect}, even for conservative values of $\tau$, is clearly identified by the widened gap after $\lambda_3$ in the eigenvalue graphs.

However, both eigenvalue plots in \fref{mccicceigs}, with and without the intolerance parameter, reveal a Perron-cluster containing $k^*=3$ eigenvalues, as desired.  

\subsubsection{Refining the Consensus Matrix through Iteration}

\begin{figure}[h!]
\begin{subfigure}[h!]{.49\linewidth}
\centering
\includegraphics[scale=.35]{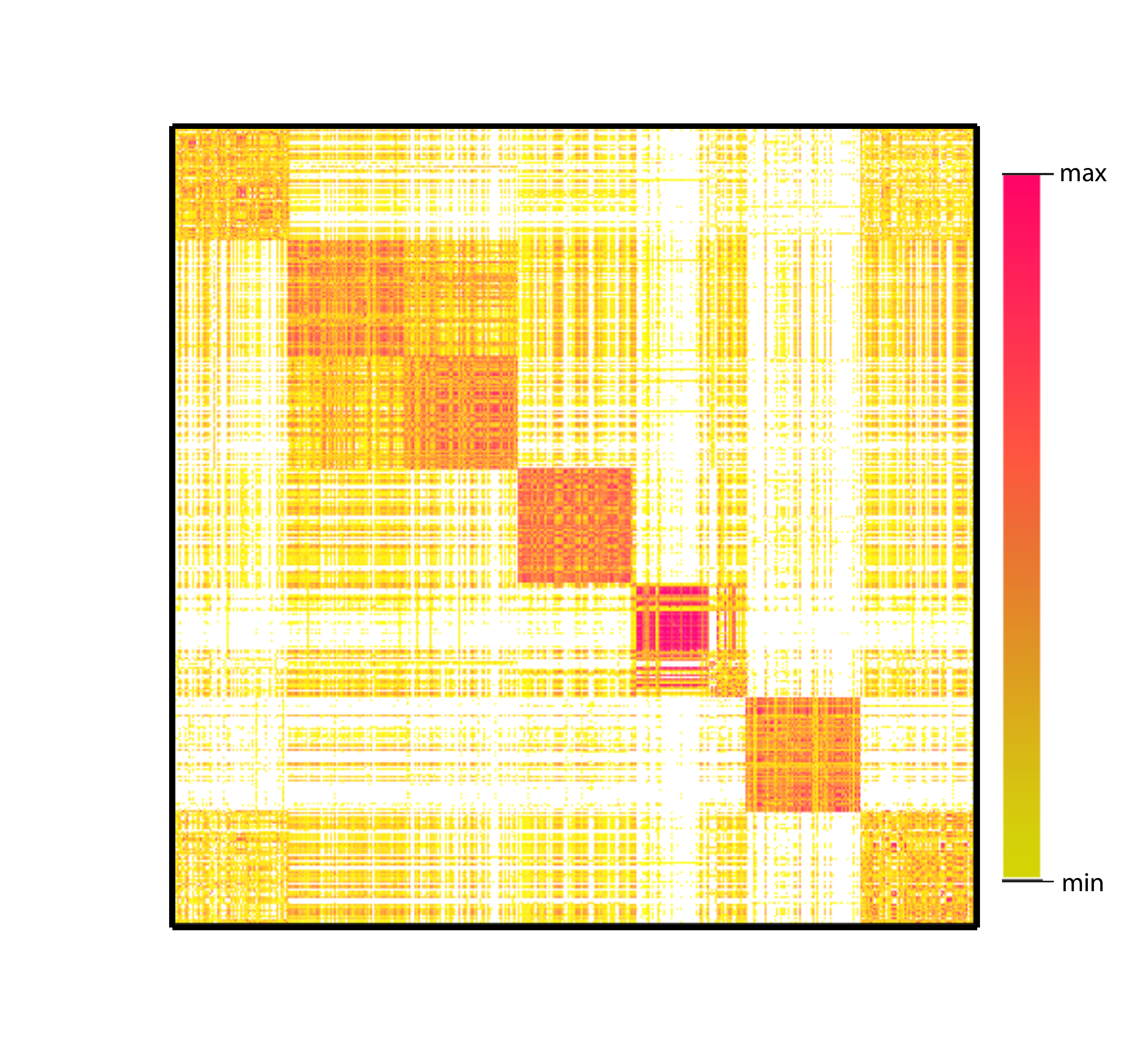}
\caption{Consensus Matrix prior to iteration}
\label{vismatrix0}
\end{subfigure}
\begin{subfigure}[h!]{.49\linewidth}
\centering
\includegraphics[scale=.35]{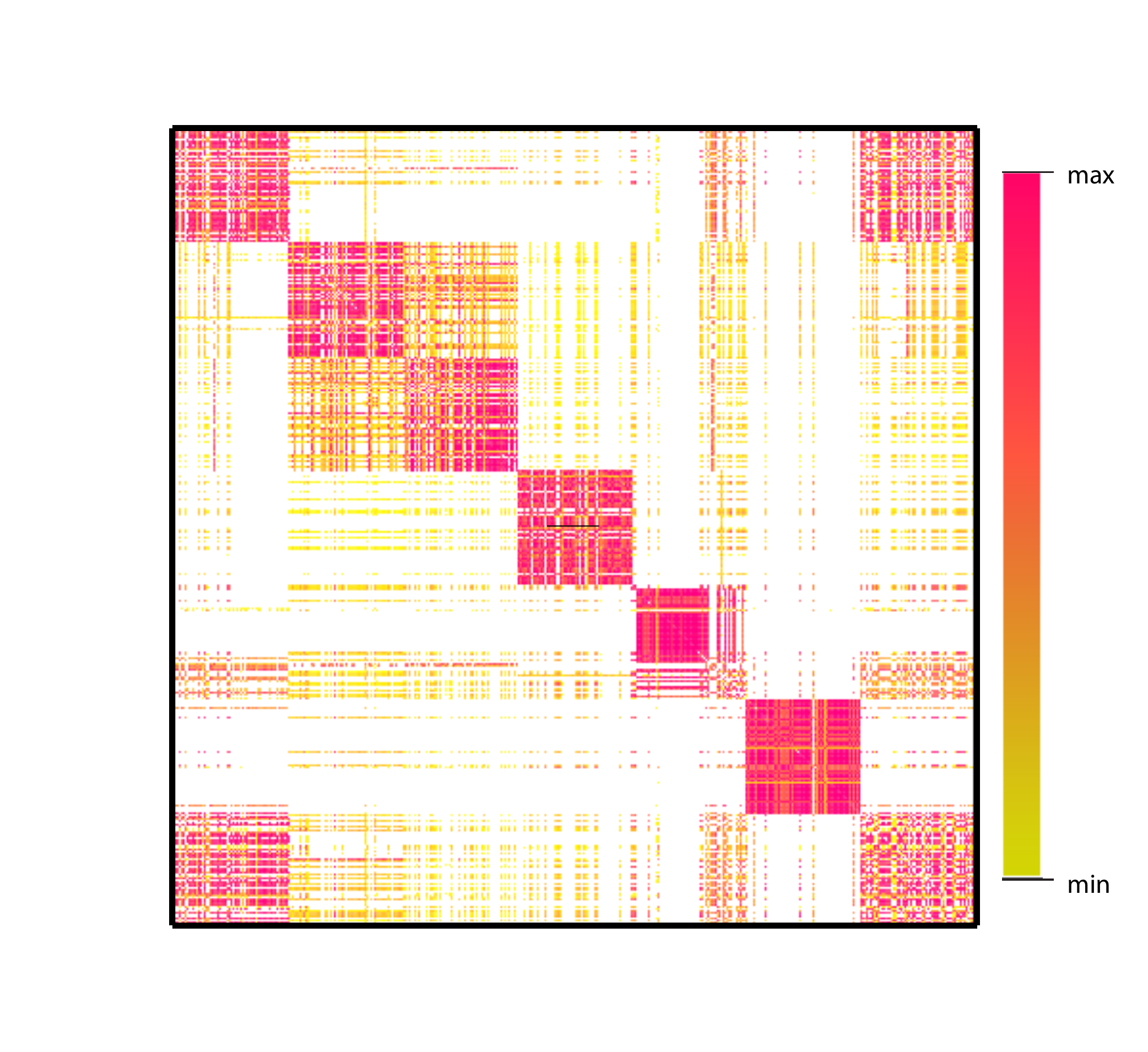}
\caption{Consensus Matrix after Iteration}
\label{vismatrix2}
\end{subfigure}
\caption{The Uncoupling Effect of Iteration: Matrix Heat Map}
\label{vismatrix}
\end{figure}

\begin{figure}[h!]
\begin{subfigure}[h]{.49\linewidth}
\centering
\includegraphics[scale=.38]{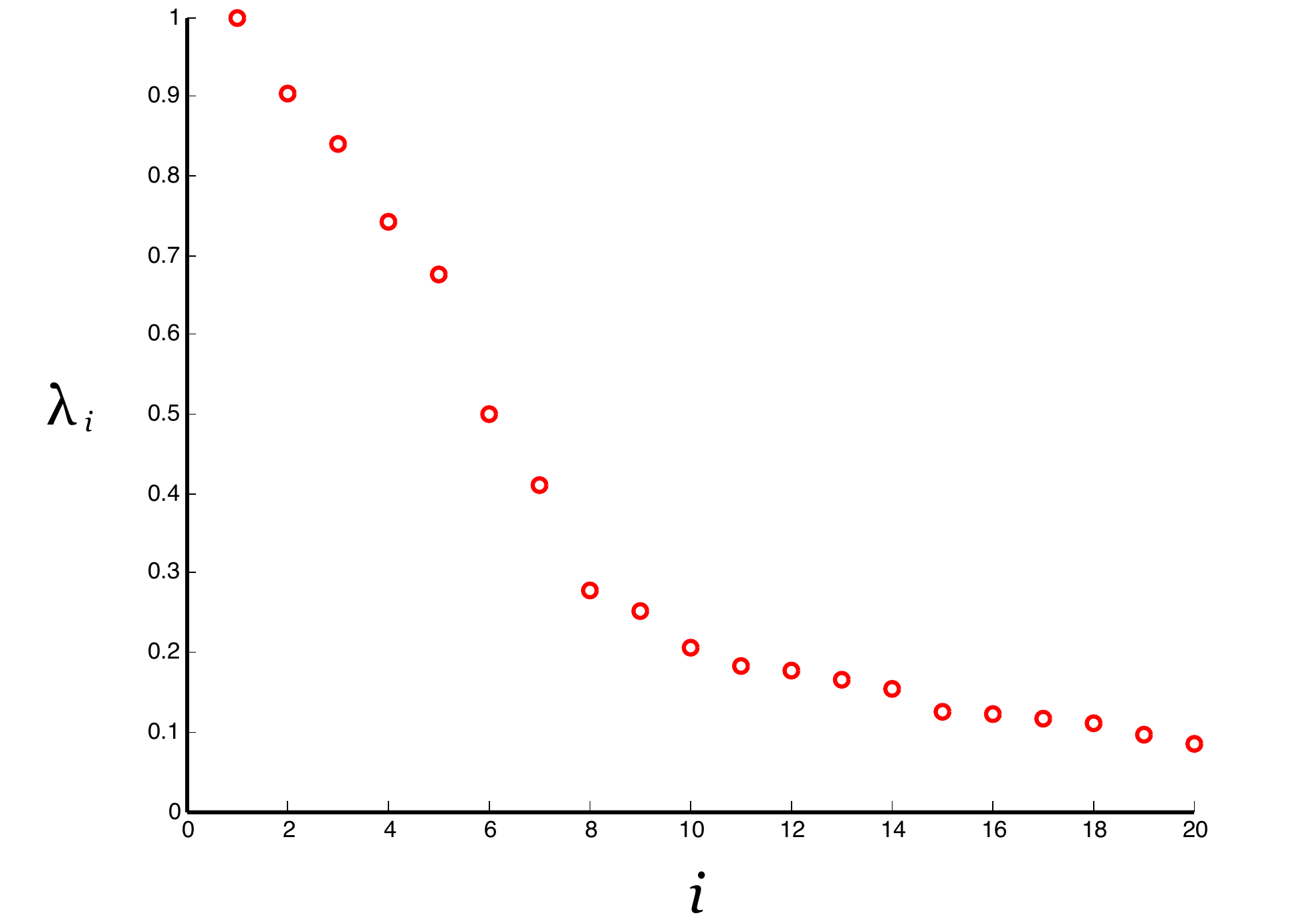}
\caption{Eigenvalues prior to iteration}
\end{subfigure}
\begin{subfigure}[h!]{.49\linewidth}
\centering
\includegraphics[scale=.38]{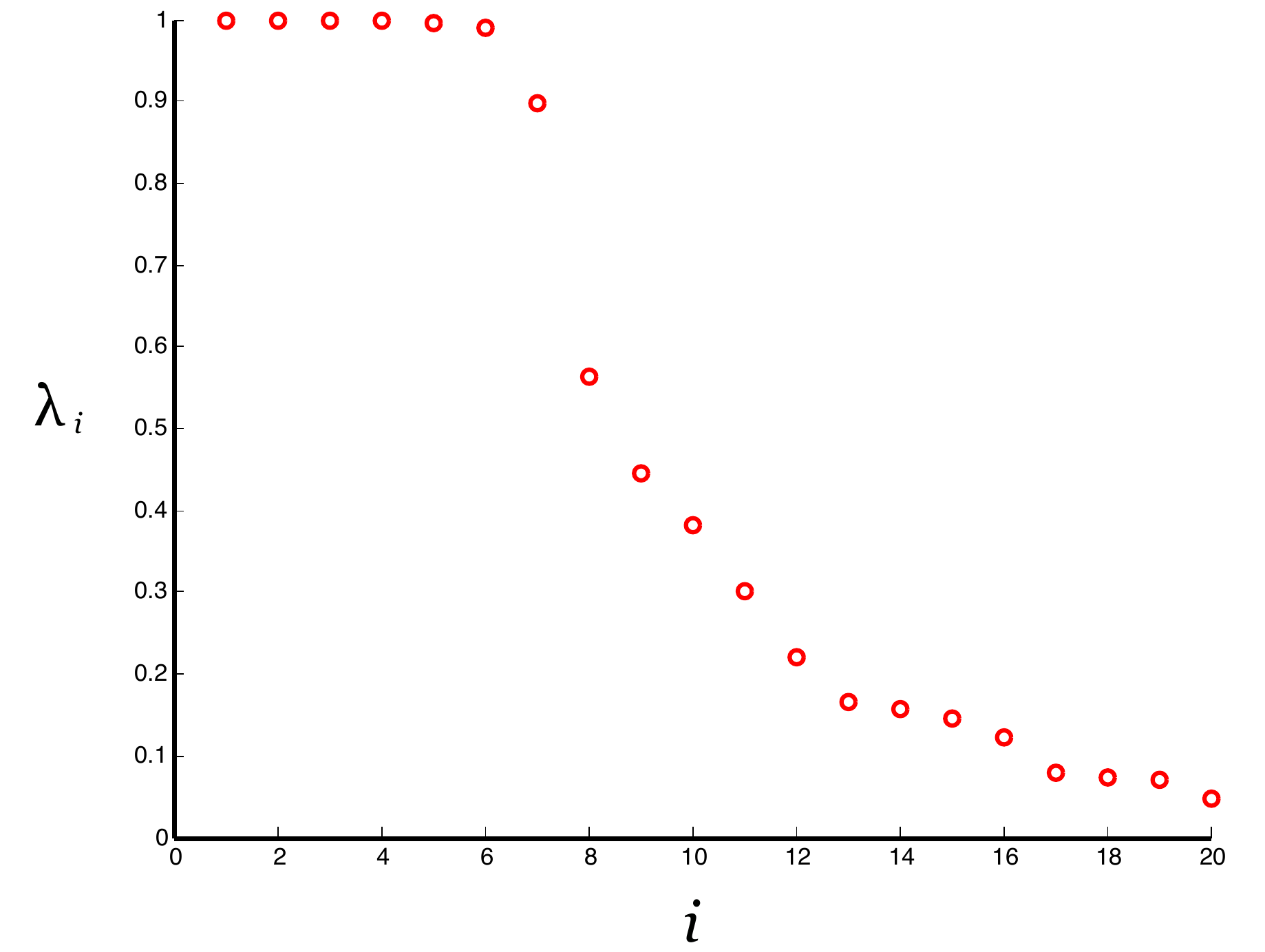}
\caption{Eigenvalues after iteration}
\end{subfigure}
\caption{The Uncoupling Effect of Iteration: Eigenvalues}
\label{viseigs}
\end{figure}

We have seen with the Medlars-Cranfield-CISI collection that the consensus matrix can provide better clustering information than the raw data. Therefore it seems reasonable that iterating the consensus process using multiple values of $\tilde{k}$ may refine the consensus matrix in way that minimizes or eliminates elements outside of the diagonal blocks, revealing a more identifiable Perron cluster. This is most often the case. Iterating the clustering process has an uncoupling effect on the consensus matrix \cite{siamicc}.  In \fref{vismatrix0} we show a matrix heat-map of a consensus matrix formed by clustering 700 documents (100 of each in 7 clusters) into $\tilde{k}=[10, 11,\dots, 20]$ clusters with 4 different algorithms and 3 different dimension reductions. Red pixels indicate high levels of similarity while yellow pixels represent lower levels of similarity. There is a considerable amount of noise outside of the diagonal blocks. This consensus matrix was then clustered by the same 4 algorithms and 3 dimension reductions, again into $\tilde{k}=[10, 11,\dots, 20]$, and a heat map of the resulting consensus matrix (iteration 2) is shown in \fref{vismatrix2}. It is easy to see the refinement of the clusterings by the reduction of noise outside the diagonal blocks. The difference is also clearly shown in the eigenvalue plots displayed in \fref{viseigs}. For high-dimensional or noise-ridden data, we suggest this iterated procedure in determining the Perron gap, because spurious cluster relationships will often couple the clusters together. 

 In \sref{results} the flexibility of our approach is demonstrated using a comprehensive example on another benchmark dataset.  The Iterative Consensus Clustering Framework is summarized in Algorithm \ref{iccframework}.  

\begin{algorithm}
\caption{Iterative Consensus Clustering (ICC) Framework}
\label{iccframework}
\textbf{Part I: Determining the Number of Clusters}
\textit{(If desired number of clusters is known, skip to Part II.)}
\begin{itemize}
\item[] \textbf{Input:} Data Matrix $\X$, intolerance parameter $\tau$ (if desired), and sequence $\tilde{k}=\tilde{k}_1,\tilde{k}_2,\dots,\tilde{k}_J$
\item[1.] Using each clustering method $i=1,\dots,N$, partition the data into $\tilde{k}_j$ clusters, $j=1,\dots,J$
\item[2.] Form a consensus matrix, $\bo{M}$ with the $JN$ different clusterings determined in step 1. 
\item[3.] Set  $\bo{M}_{ij}=0$ if $\bo{M}_{ij} <\tau JN$.
\item[4.]Let $\bo{D}=\mbox{diag} (\bo{M}\e)$. Compute the eigenvalues of $\bP$ using the symmetric matrix $\bo{I}-\bo{D}^{-1/2}\bo{M}\bo{D}^{-1/2}$.
\item[5.] Output the number of eigenvalues in the Perron cluster, $k$. Repeat steps 1-5 using $\bo{M}$ as the data input in place of $\X$, if the number of eigenvalues in the Perron cluster remains the same, stop.
\end{itemize}
\textbf{Part II: Determining the Consensus Solution}
\begin{itemize}
\item[] \textbf{Input:} Final consensus matrix from part I, intolerance parameter $\tau$  (if desired), and the number of clusters $k$. (Or if desired number of clusters is known before hand, the raw data matrix $\X$).
\item[1.] Using each clustering method $i=1,\dots,N$, partition the matrix into $\tilde{k}_j$ clusters, $j=1,\dots,J$
\item[2.] If the majority of algorithms agree upon a single solution, stop and output this solution.
\item[3.] Form a consensus matrix, $\bo{M}$ with the $JN$ different clusterings determined in step 1. 
\item[4.] Set  $\bo{M}_{ij}=0$ if $\bo{M}_{ij} <\tau JN$.
\item[5.] Repeat steps 1-5.
\end{itemize}
\end{algorithm}

\section{Comprehensive Results on a Benchmark Dataset}
\label{results}

\subsection{NG6: Subset of 20 Newsgroups}
\begin{figure}[h!]
\centering
\includegraphics[width=0.5\linewidth]{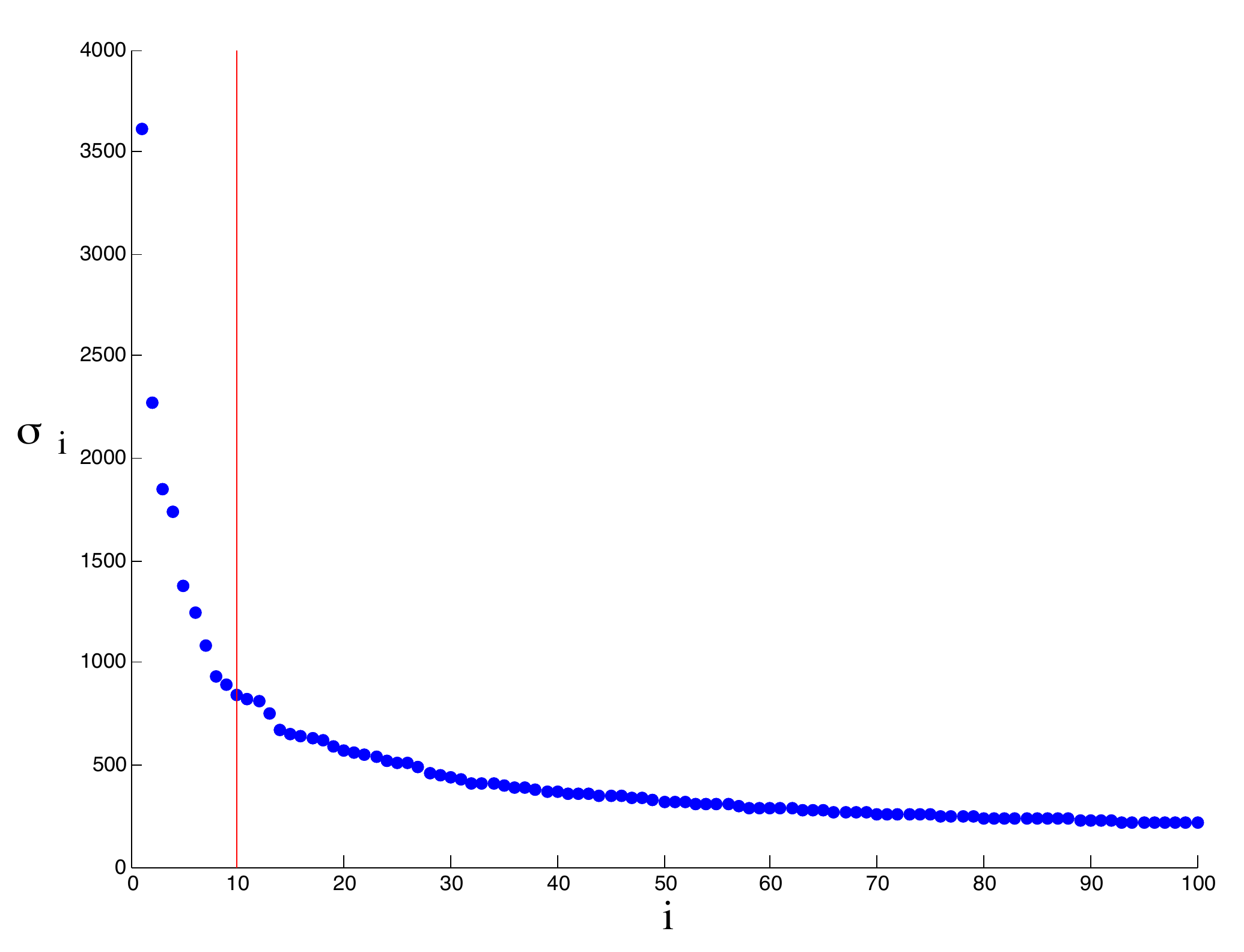}
\caption{Dataset NG6: Screeplot (First 100 Singular Values by Index)}
\label{ng6screeplot}
\end{figure}

The Newsgroups 6 (NG6) dataset is a subset from the infamous "Twenty Newsgroups" text corpus that has become a common benchmark for cluster analysis. The Twenty Newsgroups corpus consists of approximately 20,000 news documents (web articles) partitioned somewhat evenly across 20 different topics. The collection of these documents is attributed to Ken Lang, although it is never mentioned explicitly in his original paper \cite{Lang}. It is now publicly available via the web \cite{20newsgroups}.  To create the NG6 collection, 300 documents from 6 topics were randomly selected, resulting in a term-document matrix with 11,324 terms and 1800 documents. Our initial consensus matrix (used to determine the number of clusters) was formed using \textit{only} the \kmeans algorithm (randomly initialized), performed on the raw data and 3 dimension reductions. To determine an appropriate rank for dimension reduction we followed convention by observing the screeplot (plot of the singular values) for the NG6 data matrix. The screeplot shown in \fref{ng6screeplot}, indicates our decision to reduce the dimensions of the data from $m=11,324$ to $r=10$. 

The dimensionality of the data was then reduced using our 3 preferred dimension reduction algorithms:
\begin{enumerate}
\item Principal Components Analysis
\item Singular Value Decomposition
\item Nonnegative Matrix Factorization
\end{enumerate}
and 10 iterations of \kmeans clustering was performed on each dimension reduction to create $\tilde{k} = 10,11,12,\dots,20$ clusters. Since the clustering is only performed on the reduced data, this phase of the process proceeds extremely fast. The result was a total of 330 clusterings which contributed to the initial consensus matrix. An intolerance parameter was not used for this initial matrix. The next step in our analysis is to examine the eigenvalues of the transition probability matrix associated with this initial consensus matrix. 

\begin{figure}[h!]
\centering
\includegraphics[width=0.5\linewidth]{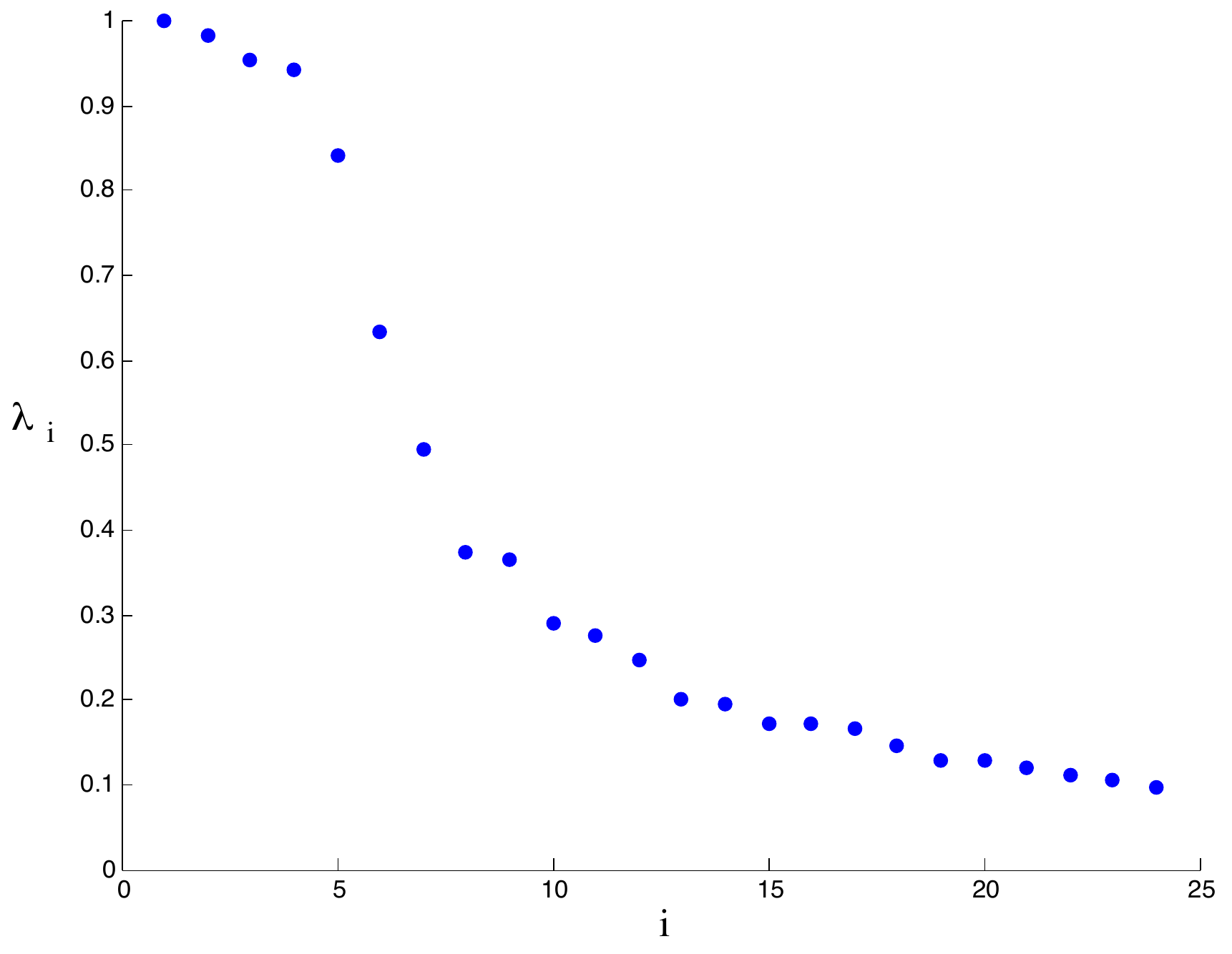}
\caption{Dataset NG6: Eigenvalues associated with Initial (unadjusted) Consensus Matrix}
\label{ng6iter0Eigs}
\end{figure}

The Perron cluster in \fref{ng6iter0Eigs} contains 5 eigenvalues for initial consensus matrix.  As discussed in \sref{perronconsensus} there are two adjustments one might consider to further explore the cluster structure.

\begin{figure}[h!]
\centering
\begin{subfigure}{.47\textwidth}
\centering
\includegraphics[width=0.9\linewidth]{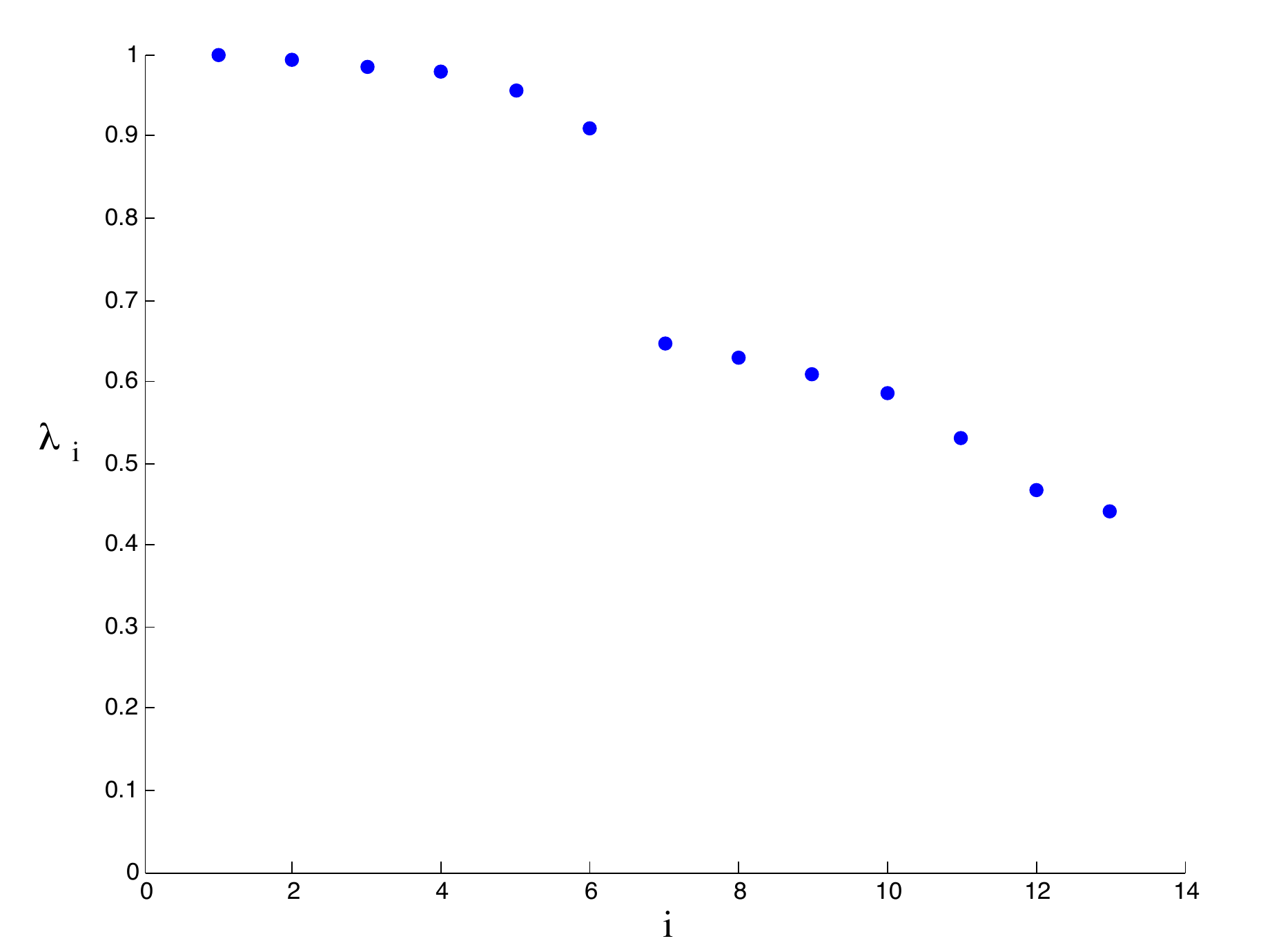}
\caption{ \centering Eigenvalues after $\tau=0.3$ \newline
			(Consensus A)}
\label{ng6eigsp30}
\end{subfigure}
\begin{subfigure}{.47\textwidth}
\centering
\includegraphics[width=0.9\linewidth]{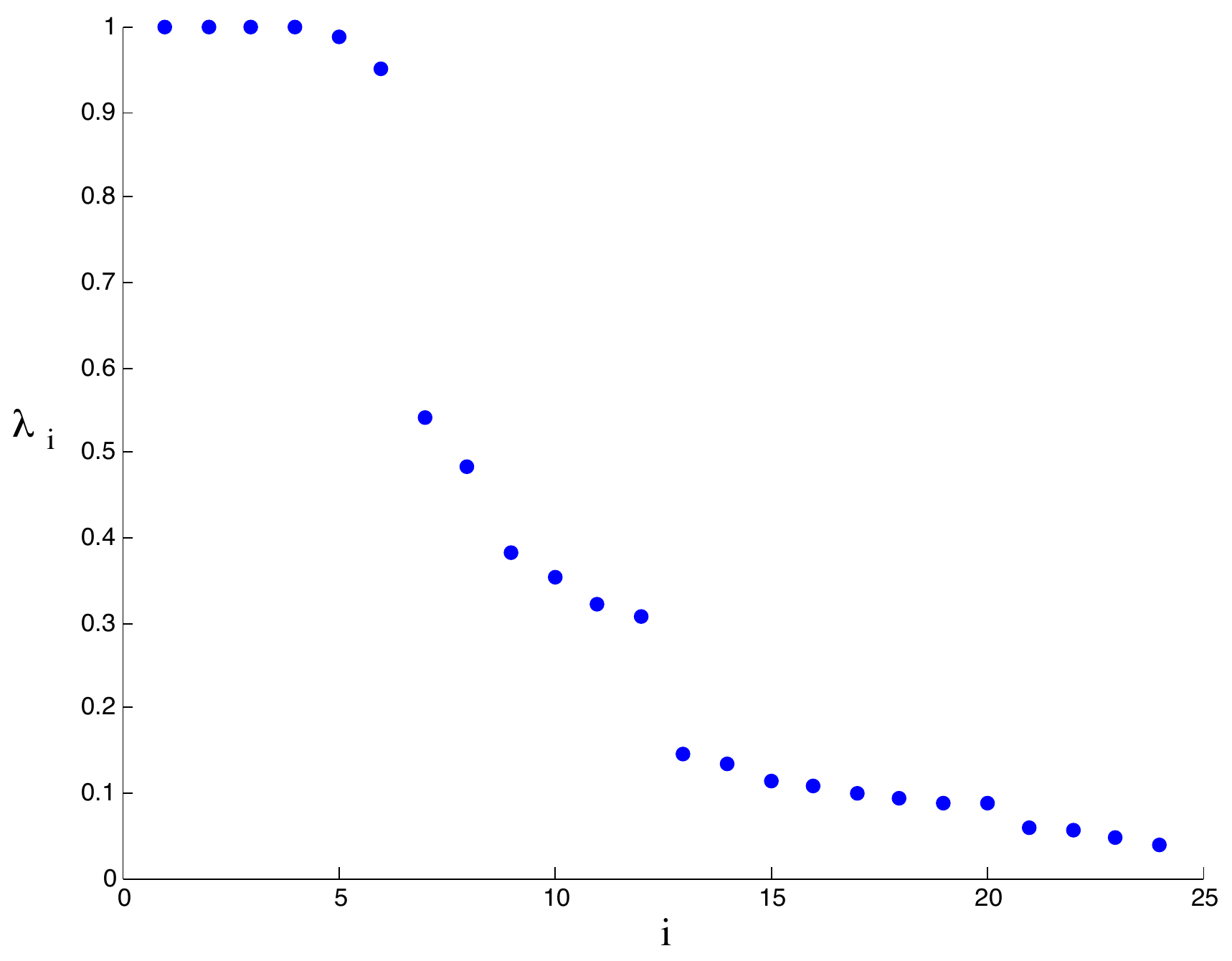}
\caption{ \centering Eigenvalues after one iteration \newline (Consensus B)}

\end{subfigure}
\caption{Dataset NG6: Eigenvalues associated with consensus similarity matrices adjusted by intolerance (right) or adjusted by iteration (left) using $\tilde{k}=10,11,12,\dots,20$ clusters. The $k^*=6$ eigenvalues in the Perron cluster correctly identify the number of clusters.}
\label{ng6adjust}
\end{figure}

\begin{enumerate}
\item Implement an intolerance parameter $\tau$ to distinguish the Perron cluster
\item Iterate the consensus procedure using the initial consensus matrix as input (no dimension reduction was used here).
\end{enumerate}
In \fref{ng6adjust} the results of both adjustments are shown. In either scenario, a Perron cluster with $k^*=6$ eigenvalues becomes clear. When the iterative procedure is repeated once more, the number of eigenvalues in the Perron cluster does not change.

\begin{figure}[h!]
\centering
\includegraphics[width=0.5\linewidth]{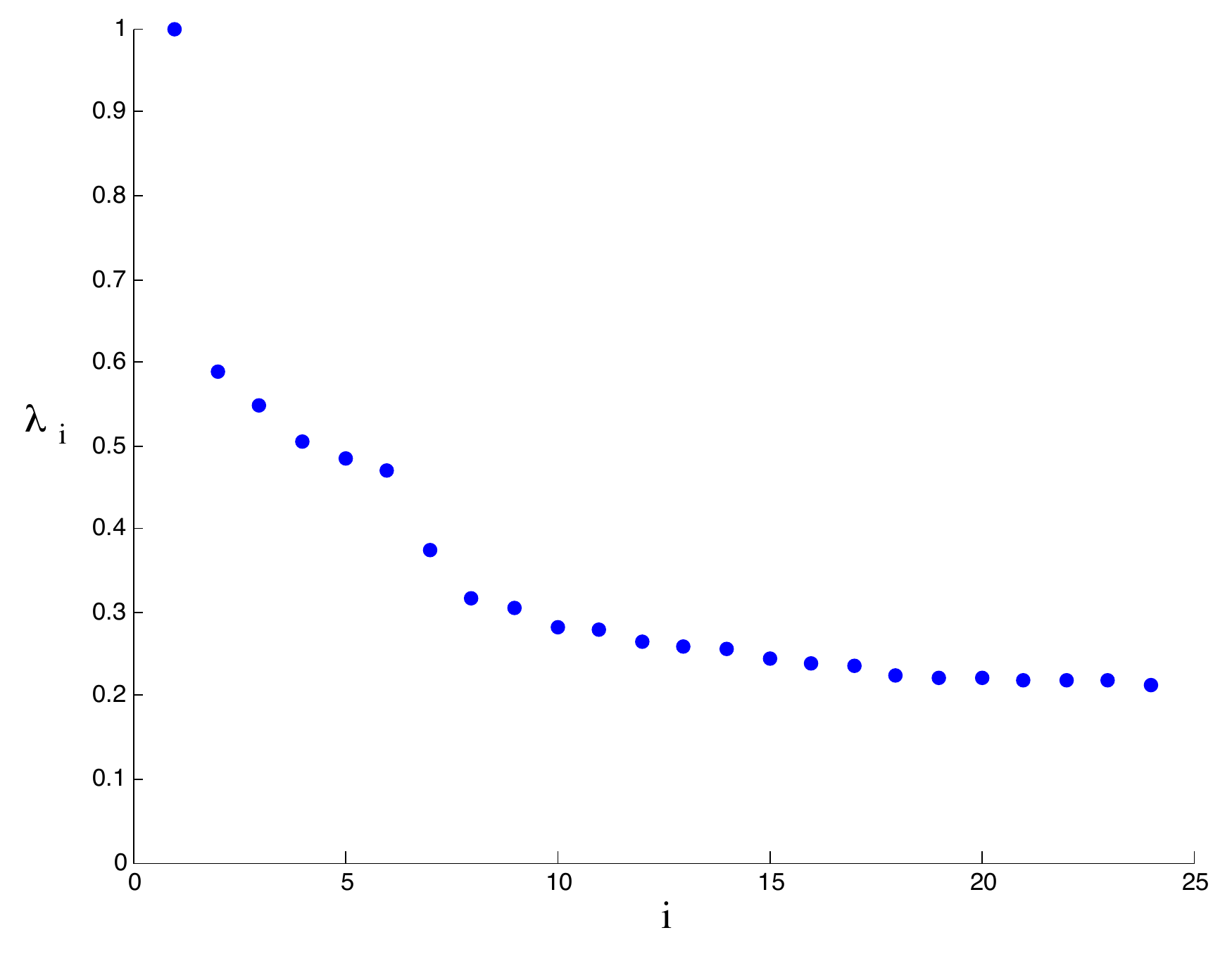}
\caption{Dataset NG6: Eigenvalues associated with Cosine Similarity Matrix}
\label{ng6cosineeigs}
\end{figure}

For the purposes of comparison, we present in \fref{ng6cosineeigs} the eigenvalues of the transition probability matrix associated with the Cosine similarity matrix, which is commonly used to cluster document datasets with spectral algorithms. No information regarding the number of clusters is revealed by the Perron cluster, which contains only a single eigenvalue.

\subsubsection{NG6 Data: Determining a Cluster Solution}
\paragraph{Comparing the Consensus Matrix as Input}
Tables \ref{ng6rawaccs}, \ref{ng6consaccs} and \ref{ng6rawaccs2} demonstrate the superiority of the consensus matrices to traditional data inputs. The accuracies of certain algorithms increase by as much as 60\% when the consensus matrix is used as input compared with the raw data. The average accuracies of all the algorithms also increases dramatically. One interesting fact to point out is that in this example, the algorithms perform quite poorly on the NMF dimension reduction. Even though these results are contained in the consensus matrices, the nature of the process is able to weed out these poor results.  The authors have experimentally discovered time and again that the ICC process is not sensitive to a small number of poor results contained in the cluster ensemble. When it comes to clustering the ensemble, such results are essentially ``voted out.'' 
\begin{table}[h!]
\centering
\begin{tabular}{c| c c c c c c c}
  Algorithm & Raw Data &  PCA $r=10$ & SVD $r=10$& NMF $r=10$ \\
 \hline
 PDDP& 0.32 & 0.75 &  0.80 & 0.31 \\
  PDDP-\kmeans & 0.77  & 0.99 & 0.99 & 0.60  \\
 
 NMF-Basic & 0.46 & - & - & - \\
 \kmeans & 0.18 &0.81 & 0.68 & 0.18 \\
   \hline
   Average & 0.43 & 0.85 & 0.82 & 0.36 
\end{tabular}
\caption{Dataset NG6: Accuracies for individual algorithms on raw data and dimension reductions.} 
\label{ng6rawaccs}
\end{table}
\begin{table}[h!]
\centering
\begin{tabular}{c| c c c c c c c}
  Algorithm  & Initial Consensus &  Consensus A & Consensus B\\
 \hline
 PDDP& 0.85 & 0.93 & 0.98\\
  PDDP-\kmeans & 0.96 & 0.94 & 0.98 \\
 
 NMF-Basic & 0.98 & 0.98 & 0.99\\
 \kmeans & 0.75 & 0.72 & 0.78\\
   \hline
   Average  & 0.88 & 0.89 & 0.93
\end{tabular}
\caption{Dataset NG6: Accuracies for individual algorithms on 3 different consensus matrices.} 
\label{ng6consaccs}
\end{table}

\begin{table}[h!]
\centering
\begin{tabular}{c| c c c c}
 Algorithm &   Cosine & Initial Consensus &  Consensus A & Consensus B\\
 \hline
  PIC &  0.52 &0.86& 0.97 & 0.98 \\
   NCUT&  0.98 & 0.98& 0.98 & 0.99\\
  NJW & 0.87 & 0.82&0.80 & 0.99\\
   \hline
   Average & 0.79 & 0.89& 0.92 & 0.99
\end{tabular}
\caption{Dataset NG6: Accuracies for spectral algorithms on different similarity matrices.} 
\label{ng6rawaccs2}
\end{table}
\subsubsection{NG6 Data: Determining a Final Solution}
Our second step in the ICC process, once the number of clusters has been determined via the eigenvalue analysis, is to iterate the consensus procedure using the determined number of clusters in an attempt to witness agreement between algorithms. Combining the clustering results of each algorithm on Consensus B (Column 3 in \tref{ng6consaccs} and Column 4 in \tref{ng6rawaccs2}) into another consensus matrix, we run through a second round of ``voting''.  The matrix Consensus B was chosen because the eigenvalue gap was larger, although using Consensus matrix A provides a similar result.  The accuracies of the resulting solutions are given in \tref{ng6consensus}. The boxed in values indicate a common solution among algorithms. We call this the \textit{final consensus solution.}

 \begin{table}[h!]
\centering
\begin{tabular}{c| c}
  Algorithm & Consensus Iter 2\\
 \hline
 PDDP& 0.83\\
  PDDP-\kmeans &  \framebox{0.99}\\
 NMF-Basic & \framebox{0.99}\\
 \kmeans & \framebox{0.99}\\
  PIC &  \framebox{0.99}\\
   NCUT& 0.99\\
  NJW &  0.82\\
 
\end{tabular}
   \caption{Dataset NG6: 4 of 7 algorithms find a common solution in one iteration of the final consensus process}
   \label{ng6consensus}
   \end{table}

\subsubsection{NG6 Data: Conclusion}

We began our analysis with a collection of documents and algorithms - both for dimension reduction and for clustering. Traditional tools for determining the number of clusters were not successful. An analyst, having somehow determined the number of clusters and attempting to cluster this data with the given set of tools had a chance of finding a solution with accuracy ranging from 18\% to 99\%. If that analyst had chosen the best dimension reduction algorithm (PCA) for this particular dataset (a task for which there are no hard and fast guidelines), the accuracy of his/her solution may have been between 77\% and 99\%. Internal cluster validation metrics like the \kmeans objective function would not have been much help in choosing between these solutions, as such measures are difficult to compare on high-dimensional data. However, by using \textit{all} of the tools at our disposal in the Iterative Consensus Clustering Framework, we found that the clustering algorithms worked out their differences constructively - finally settling down on a solution with the highest level of accuracy achieved by any of the algorithms independently.  

\section{Conclusion}
Herein we have presented a flexible framework for combining results from multiple clustering algorithms and/or multiple data inputs. Not only does this framework provide the user with an above average clustering solution, it also contains a practical exploratory procedure for determining the number of clusters.

We have discovered that consensus matrices built using multiple algorithms and multiple values for the number, $k$, of clusters will often allow users to estimate an appropriate number of clusters in data by determining the maximum number of clusters for which algorithms are likely to agree on a common solution.
We have provided several examples to show how this approach succeeds at determining the number of clusters in datasets where other methods fail. When the initial consensus matrix does not provide this information, it can be refined through the use of an intolerance parameter or iteration to get a clearer picture of how many clusters the algorithms might be able to agree upon. 

While the consensus matrix itself is not a new idea, the practice of using multiple algorithms and dimension reductions together to create the matrix had not previously been explored, nor had varying the number of clusters for the purposes of approximating $k$. Our approach to building the consensus matrix is novel and improves clustering results from nearly every clustering algorithm on all datasets considered. This consensus matrix has several advantages over traditional similarity matrices as discussed in \sref{benefits}.

The ICC Framework encourages clustering algorithms to agree on a common solution to help escape the unreliability of individual algorithms. While previous consensus methods have aimed to average cluster solutions in one way or another, ours is the first to emphasize agreement between clustering algorithms.  After seeing some of the results of the individual algorithms in our ensemble, it should be clear that an average solution could be very poor indeed. Rather than deciding each clustering is equally valid, we simply sum the number of times a cluster relationship was made between two points and let the algorithms decide whether this sum is considerable enough to draw those points together, or whether it might be more reasonable to dissolve the connection in favor of others. This framework iteratively encourages algorithms to agree upon a common solution because the value of the similarity metric reflects the level of algorithmic agreement at each step. Thus, through iteration, cluster relationships upon which the algorithms do not agree are abandoned in favor of relationships with higher levels of agreement.

\end{document}